\documentclass{article}

\usepackage{arxiv}

\usepackage[utf8]{inputenc} 
\usepackage[T1]{fontenc}    
\usepackage{hyperref}       
\usepackage{url}            
\usepackage{booktabs}       
\usepackage{amsfonts}       
\usepackage{amsmath}        
\usepackage{nicefrac}       
\usepackage{microtype}      
\usepackage{lipsum}		
\usepackage{graphicx}
\usepackage{natbib}
\usepackage{doi}
\usepackage{algorithm}
\usepackage{algorithmic}
\usepackage{longtable}
\usepackage{float} 
\usepackage{tikz}

\title{VSCOUT: A Hybrid Variational Autoencoder Approach to Outlier Detection in High-Dimensional Retrospective Monitoring}

\date{} 					

\author{ \href{https://orcid.org/0000-0002-6634-9016}{\includegraphics[scale=0.06]{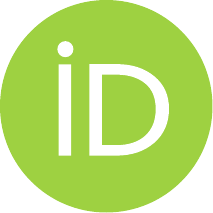}\hspace{1mm}Waldyn Martinez} \\
	Department of Information Systems \& Analytics\\
	Miami University\\
	Oxford, OH 45056 \\
	\texttt{martinwg@miamioh.edu} \\
	  \\
}

\renewcommand{\shorttitle}{VSCOUT: A Hybrid Variational AutoEncoder Approach to Outlier Detection}

\hypersetup{
pdftitle={A template for the arxiv style},
pdfsubject={q-bio.NC, q-bio.QM},
pdfauthor={David S.~Hippocampus, Elias D.~Striatum},
pdfkeywords={First keyword, Second keyword, More},
}

\begin{document}
\maketitle

\begin{abstract}
Modern industrial and service processes generate high-dimensional, non-Gaussian, and contamination-prone data that challenge the foundational assumptions of classical Statistical Process Control (SPC). Heavy tails, multimodality, nonlinear dependencies, and sparse special-cause observations can distort baseline estimation, mask true anomalies, and prevent reliable identification of an in-control (IC) reference set. To address these challenges, we introduce \textbf{VSCOUT}, a distribution-free framework designed specifically for \emph{retrospective} (Phase~I) monitoring in high-dimensional settings. VSCOUT combines an Automatic Relevance Determination Variational Autoencoder (ARD-VAE) architecture with ensemble-based latent outlier filtering and changepoint detection. The ARD prior isolates the most informative latent dimensions, while the ensemble and changepoint filters identify pointwise and structural contamination within the determined latent space. A second-stage retraining step removes flagged observations and re-estimates the latent structure using only the retained inliers, mitigating masking and stabilizing the IC latent manifold. This two-stage refinement produces a clean and reliable IC baseline suitable for subsequent Phase~II deployment. Extensive experiments across benchmark datasets demonstrate that VSCOUT achieves superior sensitivity to special-cause structure while maintaining controlled false alarms, outperforming classical SPC procedures, robust estimators, and modern machine-learning baselines. Its scalability, distributional flexibility, and resilience to complex contamination patterns position VSCOUT as a practical and effective method for retrospective modeling and anomaly detection in AI-enabled  environments.
\end{abstract}


\section{Introduction}

Recent advances in sensing and data acquisition have led to the widespread
availability of high-dimensional process data, characterized by nonlinear
dynamics and complex dependence structures that strain the assumptions of
classical Statistical Process Control (SPC). Within such environments,
retrospective (Phase~I) analysis remains essential for constructing a reliable
in-control (IC) reference before any prospective monitoring can be meaningfully
performed. Yet Phase~I data are often contaminated by unknown special-cause
observations, transient disruptions, heavy tails, multimodality, or
nonstationarity, all of which can distort baseline estimates and undermine
downstream control charts
\citep{reis2017industrial, montgomery2020introduction, goecks2024industry}.

Traditional multivariate approaches such as Hotelling's $T^2$ chart \citep{hotelling1931generalization} and PCA-based charts \citep{ranger1996choosing, mastrangelo1996statistical} rely on linearity and multivariate normality assumptions. Even kernel-based or robust subspace variants \citep{sun2003kernel, camci2008robust} face difficulties when data exhibit heavy-tailed behavior, localized contamination, or complex latent structure. These issues are amplified in Phase~I, where the primary objective is \emph{to identify and remove contaminated observations} before estimating IC parameters. When contamination is present, classical estimators become biased and masking effects degrade the latent or covariance structures upon which Phase~II monitoring depends.

Deep generative models provide a flexible alternative for modeling high-dimensional data, with variational autoencoders (VAEs) \citep{kingma2013auto} offering a principled way to learn nonlinear latent structure. However, standard VAEs are not tailored to Phase~I objectives: they may blur the separation between inliers and outliers, yield unstable reconstruction-based thresholds, or overfit anomalous samples when trained directly on unfiltered retrospective data \citep{ruff2021unifying}. Because VAEs aim to maximize generative likelihood rather than detect anomalies, their latent spaces can misrepresent common-cause variation when contamination is present.

To address these challenges, we propose VSCOUT (VAE Self-Correcting Outlier Uncovering Technique), a hybrid deep-learning framework designed specifically for high-dimensional retrospective monitoring. VSCOUT trains a VAE to learn an initial latent structure, applies ensemble-based detectors and changepoint analysis to identify potential special-cause observations, removes these points, and then \emph{refines} the VAE on the cleaned dataset. Instead of restarting training, the refinement stage continues optimizing the encoder–decoder weights from the initial fit, preserving useful structure while reducing the influence of anomalous data. The resulting latent representation more closely aligns with an ideal Gaussian prior, providing stable thresholds and improved separation between IC and OC observations.

We evaluate VSCOUT on extensive simulations that reflect diverse Phase~I contamination scenarios, including heavy-tailed, log-normal, mixed, and multimodal distributions, and on widely used real-world benchmark datasets. Across all experiments, VSCOUT reliably identifies clean inlier subsets, recovers stable IC latent representations, and outperforms leading SPC, ensemble, and deep anomaly detection baselines in detecting special-cause structure while controlling false alarms.

Our contributions are threefold:
\begin{itemize}
    \item We introduce VSCOUT, a distribution-free retrospective (Phase~I) framework that integrates ARD-VAE modeling, ensemble filtering, and changepoint detection to construct robust IC baselines for high-dimensional processes.
    \item We provide a comprehensive benchmarking study comparing VSCOUT to classical SPC tools, robust estimators, and state-of-the-art anomaly detection models across synthetic and real datasets.
    \item We demonstrate that VSCOUT achieves superior sensitivity to special-cause structure and high inlier retention even under heavy-tailed noise, nonlinear dependencies, and low signal-to-noise regimes.
\end{itemize}

The remainder of the paper is organized as follows. Section~\ref{sec:related} reviews related work in SPC and anomaly detection. Section~\ref{sec:methodogy} presents the VSCOUT framework. Section~\ref{sec:experiments} describes the simulation design and benchmarking results. Section~\ref{sec:conclusion} concludes with implications and future directions.

\section{Preliminaries}
\label{sec:related}

This section reviews the foundational concepts that underpin VSCOUT, including classical and modern approaches to Statistical Process Control (SPC), machine learning–based outlier detection, and the deep generative modeling principles relevant to variational autoencoders. We first outline traditional and contemporary methods for retrospective analysis and online monitoring, and then introduce the preliminaries of VAEs that motivate the design of our framework.

\subsection{Statistical Process Control}

For nearly a century, Statistical Process Control (SPC) has served as a foundational methodology for quality engineering, providing tools to distinguish between inherent common-cause variation and special-cause (assignable) variation that signals a shift or degradation in the process \citep{woodall1999research, montgomery2020}. SPC is conventionally organized into two stages: a retrospective analysis stage and an online monitoring stage. Although this two-stage structure remains standard, more recent research has explored unified or adaptive approaches that blur the distinction between phases in favor of integrated, data-driven monitoring frameworks \citep{hou2020new, bourazas2022predictive}.

\subsubsection{Retrospective Analysis and Online Monitoring}

Retrospective analysis refers to the examination of historical process data with the objective of establishing a reliable in-control (IC) baseline. This stage—traditionally known as Phase~I and increasingly referred to as \emph{retrospective monitoring}—is essential for characterizing natural process variability, identifying and removing assignable causes, and estimating parameters that will later define control limits. In practice, Phase~I datasets rarely include labels indicating out-of-control (OC) states, and they often contain a mix of transient disturbances, sustained shifts, and distributional irregularities. These characteristics motivate the development of unsupervised, distribution-free, and contamination-robust approaches for identifying IC regions before any online charting can be meaningfully performed \citep{qiu2017machine}.

A variety of methods address the challenges of retrospective monitoring. Changepoint detection techniques provide a principled means of segmenting historical data into homogeneous IC regions \citep{hawkins2003changepoint, zamba2006multivariate, qiu2013introduction}. Nonparametric and rank-based charts avoid strong distributional assumptions \citep{chen2016distribution}, while Distribution-Free procedures offer robustness to heavy tails and model misspecification through resampling strategies \citep{capizzi2017phase}. More recently, deep generative models, including variational autoencoders, have been used to uncover latent structure in high-dimensional retrospective datasets and to support unsupervised identification of anomalous or contaminated observations \citep{kingma2013auto}.

In contrast, online monitoring (Phase~II) applies the IC model obtained from retrospective analysis to incoming data, typically using Shewhart, CUSUM, EWMA, or similar control charts \citep{qiu2013introduction, montgomery2020}. Although modern research has explored machine-learning–based and adaptive extensions for prospective monitoring, these methods rely fundamentally on the quality of the IC baseline established during retrospective analysis. For this reason, the present work focuses on retrospective (Phase~I) modeling: VSCOUT is designed to identify and remove contamination, extract stable latent structure, and produce a reliable IC reference that can subsequently be used for Phase~II monitoring if desired.

\subsection{Outlier and Anomaly Detection Methodologies}

Anomaly detection, also referred to as outlier detection, is the task of identifying observations or patterns that deviate markedly from expected behavior. Within the context of Statistical Process Control (SPC), anomaly detection plays a central role in identifying departures from the in-control (IC) state that may indicate process changes, sensor faults, or other assignable causes \citep{chandola2009anomaly, hodge2004survey}. Anomalies typically manifest as \textit{point anomalies} (isolated extreme observations), \textit{contextual anomalies} (observations that are anomalous only under specific operating regimes such as time, load, or temperature), or \textit{collective anomalies} (groups or sequences of observations that appear anomalous only when considered jointly) \citep{chandola2009anomaly}. In retrospective SPC analysis, point anomalies often correspond to transient disruptions, contextual anomalies align with regime-specific behavior, and collective anomalies reflect sustained shifts or drifts that traditional parametric charts frequently miss. Consequently, anomaly detection provides a rich set of tools for separating reliable IC data from corrupted segments prior to establishing baselines for online monitoring.

A wide range of methodologies has been developed in the anomaly detection literature. Distance-based methods such as $k$-Nearest Neighbors (KNN) detect anomalies by identifying points that reside in sparse regions of the feature space \citep{ramaswamy2000efficient}. Density-based approaches like Local Outlier Factor (LOF) compare local density estimates to detect points in regions of relative sparsity \citep{breunig2000lof}. Ensemble-based techniques such as Isolation Forest (IForest) use random partitioning to isolate anomalous observations efficiently, making them effective in high-dimensional environments \citep{liu2008isolation}. More recently, distribution-free methods such as Empirical Cumulative Distribution Outlier Detection (ECOD) have gained prominence due to their scalability and robustness in heterogeneous, nonparametric settings \citep{li2022ecod}.

Deep learning has further expanded the scope of anomaly detection through generative and representation-learning paradigms. Variational Autoencoders (VAEs), for instance, learn nonlinear latent representations of process data, with anomalies manifesting as poor reconstructions or atypical latent embeddings \citep{kingma2013auto}. Such models are particularly valuable in high-dimensional SPC applications, where they serve both as feature extractors and as anomaly detectors, enabling the detection of subtle or structured deviations that classical models struggle to capture. Importantly, modern systems increasingly adopt ensemble strategies that integrate statistical, machine learning, and deep generative methods to exploit complementary strengths and enhance robustness across diverse data regimes.

Overall, the evolution of anomaly detection—from classical statistical approaches to advanced representation-learning methods—closely aligns with the objectives of modern SPC: robust retrospective analysis to establish reliable in-control baselines and adaptive online monitoring capable of identifying emerging process changes in real time. This broad methodological toolbox provides the foundation upon which hybrid frameworks such as VSCOUT are built.

\subsection{Deep Generative Models for Anomaly Detection in SPC}

While classical anomaly detection techniques provide valuable tools for SPC, they often face limitations in capturing the nonlinear dependencies and high-dimensional structures that characterize modern process data. Deep learning, and in particular deep generative models, have emerged as powerful alternatives capable of learning flexible representations of normal process behavior in an unsupervised or semi-supervised fashion \citep{lee2024robust}. Trained primarily on in-control (IC) data, such models approximate the underlying distribution of stable operations, enabling deviations from this learned structure to be flagged as anomalies. This modeling approach is naturally suited for adaptive online monitoring and retrospective analysis in complex, data-rich SPC environments.

\begin{figure}[ht]
  \centering
  \includegraphics[width=0.5\linewidth]{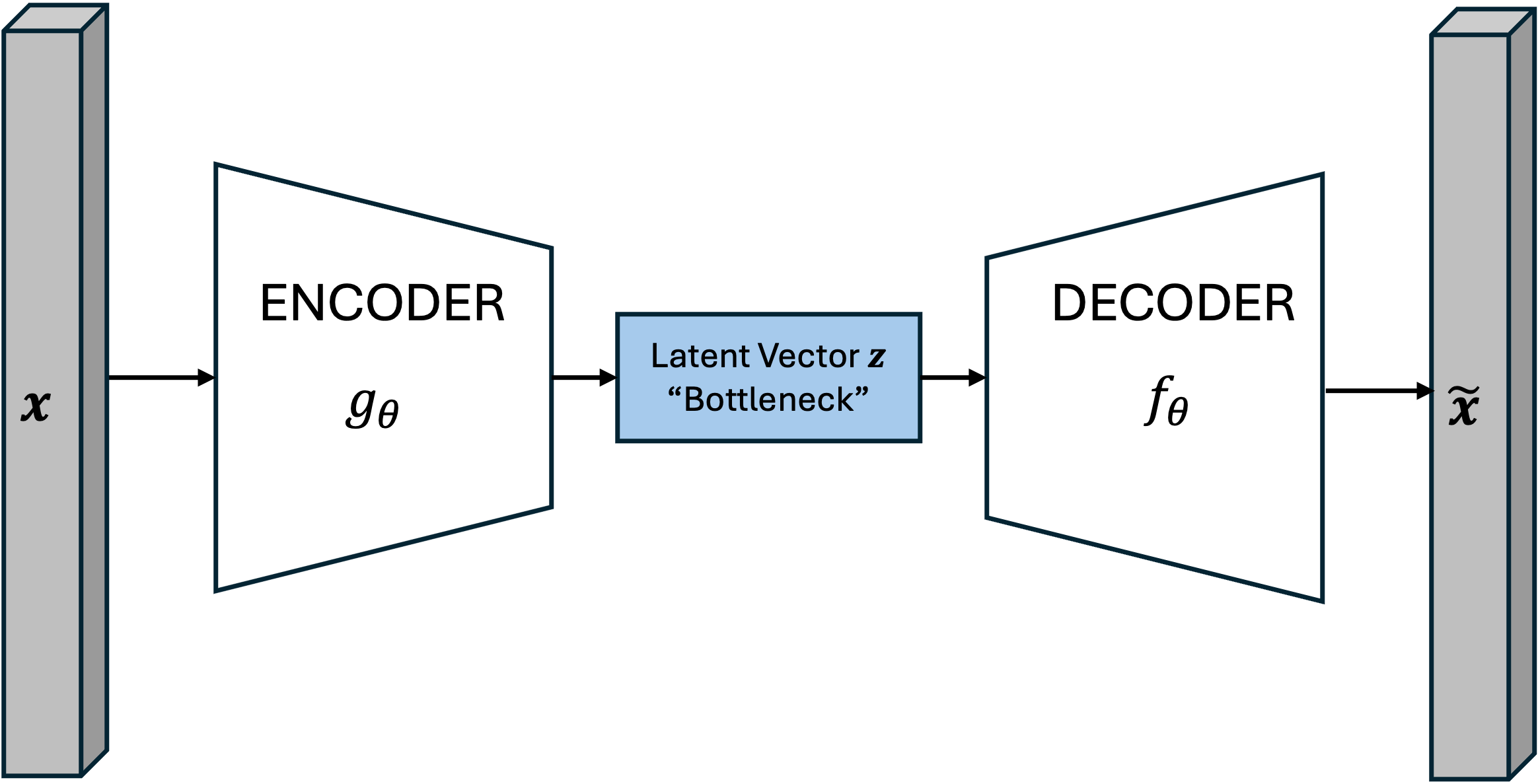}
  \caption{Autoencoder architecture.}
  \label{fig:autoencoder}
\end{figure}

A foundational deep-learning model for anomaly detection is the \textbf{autoencoder (AE)} (Figure~\ref{fig:autoencoder}). An AE consists of an encoder  $g_\theta: \mathbb{R}^p \rightarrow \mathbb{R}^d$ that maps a $p$-dimensional input $\mathbf{x}$ to a latent representation $\mathbf{z} = g_\theta(\mathbf{x})$, and a decoder $f_\theta: \mathbb{R}^d \rightarrow \mathbb{R}^p$ that reconstructs the input as  $\tilde{\mathbf{x}} = f_\theta(\mathbf{z})$. The parameters $\theta$ are learned by minimizing a reconstruction loss, typically the mean squared error (MSE):
\[
\mathcal{L}_{\text{AE}}(\theta) 
= \frac{1}{n} \sum_{i=1}^{n} 
\left\| \mathbf{x}_i - f_\theta(g_\theta(\mathbf{x}_i)) \right\|_2^2.
\]

Because the AE learns a compact nonlinear manifold capturing the structure of normal IC data, the reconstruction quality provides a natural anomaly indicator. For each observation $\mathbf{x}_i$, the \textit{reconstruction error} is defined as
\[
r(\mathbf{x}_i) 
= \left\| \mathbf{x}_i - \tilde{\mathbf{x}}_i \right\|_2^2
= \left\| \mathbf{x}_i - f_\theta(g_\theta(\mathbf{x}_i)) \right\|_2^2.
\]
Under IC conditions, $r(\mathbf{x}_i)$ tends to be small for each observation because the model has learned to reproduce nominal patterns. Out-of-control (OC) observations lie off the learned manifold and therefore yield significantly larger values of $r(\mathbf{x}_i)$. This mapping from reconstruction error to anomaly score makes the AE directly compatible with SPC-style monitoring through thresholding or control limits. Training of AEs relies on gradient-based optimization—typically stochastic gradient descent (SGD) \citep{amari1993backpropagation} or Adam \citep{kingma2014adam}—in which the model parameters $\theta$ are iteratively updated via
\[
\theta \leftarrow \theta - \eta \, \nabla_\theta \mathcal{L}_{\text{AE}}(\theta),
\]
where $\eta$ denotes the learning rate.

\begin{figure}[ht]
  \centering
  \includegraphics[width=0.6\linewidth]{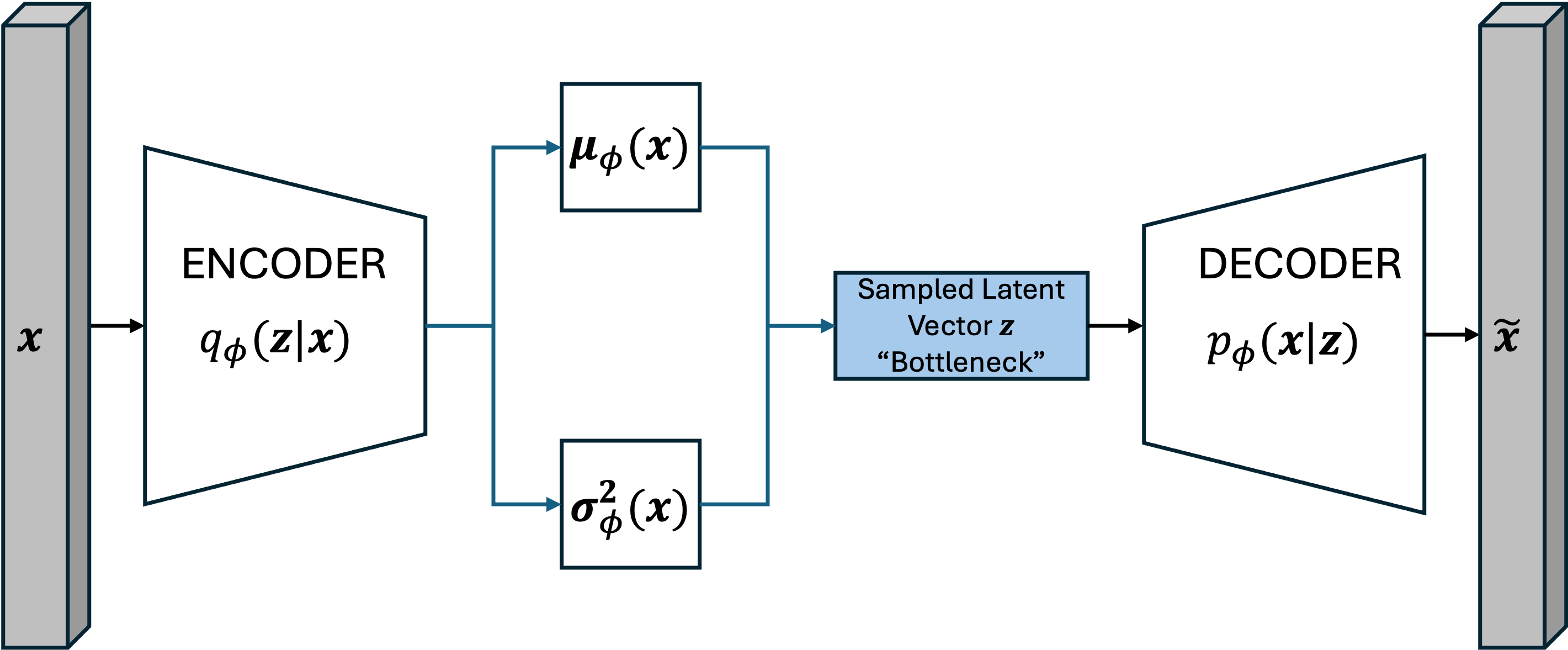}
  \caption{Variational Autoencoder (VAE) architecture.}
  \label{fig:variationalautoencoder}
\end{figure}

The \textbf{Variational Autoencoder (VAE)} extends the AE by introducing a probabilistic latent space (see Figure~\ref{fig:variationalautoencoder}). Instead of mapping inputs to a single point, the encoder outputs the parameters of an approximate posterior distribution $q_\phi(\mathbf{z} \mid \mathbf{x}) 
= \mathcal{N}\big(\boldsymbol{\mu}_\phi(\mathbf{x}), \; 
\operatorname{diag}(\boldsymbol{\sigma}^2_\phi(\mathbf{x})) \big),
$
where $\phi$ denotes encoder parameters and  
$\boldsymbol{\mu}_\phi(\mathbf{x}), \boldsymbol{\sigma}^2_\phi(\mathbf{x}) \in \mathbb{R}^d$. Latent samples are obtained via the reparameterization trick:
$
\mathbf{z} 
= \boldsymbol{\mu}_\phi(\mathbf{x}) 
+ \boldsymbol{\sigma}_\phi(\mathbf{x}) \odot \boldsymbol{\epsilon}$, where
$\boldsymbol{\epsilon} \sim \mathcal{N}(\mathbf{0}, \mathbf{I}),
$
which maintains gradient flow through stochastic sampling.

The decoder defines a likelihood model $p_\theta(\mathbf{x} \mid \mathbf{z})$, often Gaussian with fixed variance, generating a reconstruction $\tilde{\mathbf{x}}$ from $\mathbf{z}$. The VAE is trained by maximizing the evidence lower bound (ELBO):
\[
\mathcal{L}_{\text{ELBO}}(\theta, \phi; \mathbf{x})
=
\underbrace{
\mathbb{E}_{q_\phi(\mathbf{z}\mid\mathbf{x})}
\big[\log p_\theta(\mathbf{x}\mid\mathbf{z})\big]
}_{\text{Reconstruction (likelihood) term}}
-
\beta \,
\underbrace{
D_{\mathrm{KL}}\!\left(
q_\phi(\mathbf{z}\mid\mathbf{x}) \,\|\, p(\mathbf{z})
\right)
}_{\text{Regularization term}},
\]
where $p(\mathbf{z}) = \mathcal{N}(\mathbf{0}, \mathbf{I})$ is the usual prior and $\beta$ controls the strength of latent regularization. Gradients of the ELBO with respect to $(\theta, \phi)$ are computed via backpropagation:
\[
(\theta, \phi) \leftarrow (\theta, \phi) - \eta \, \nabla_{(\theta,\phi)} \mathcal{L}_{\text{ELBO}}.
\]

The probabilistic latent structure of the VAE enables both reconstruction-based and density-based anomaly detection, making it well suited for SPC applications requiring flexible modeling of complex process distributions. However, VAEs can suffer from posterior collapse, sensitivity to prior misspecification, and difficulty enforcing Gaussian structure in the latent space \citep{lin2019balancing}, especially under contamination. These limitations motivate hybrid deep generative approaches—such as VSCOUT—that combine reconstruction errors, latent-space deviation measures, and ensemble anomaly scoring to improve robustness in both Phase~I retrospective analysis and Phase~II online monitoring.

Despite their advantages, AEs and VAEs also exhibit notable limitations when the fraction of outliers in the training data becomes large. Because the reconstruction loss directly drives parameter updates, contaminated observations can disproportionately influence the gradient, pulling the learned manifold toward anomalous regions. This effect leads to manifold distortion , in which the model partially reconstructs outliers rather than separating them. As the proportion of OC data increases, the reconstruction error gap between IC and OC observations narrows, reducing discriminative power. VAEs are similarly affected: their latent distributions become biased toward contaminated regions \citep{akrami2019robust_vae, li2020rethink}, degrading the Gaussian structure of the latent space and causing the KL-regularization term to misrepresent the true IC variability. 

\section{Methodology}
\label{sec:methodogy}

The inherent susceptibility of VAEs to data contamination necessitates a specialized training regime for Statistical Process Control (SPC). While the previous section detailed how out-of-control (OC) observations distort the latent manifold, the practical challenge in retrospective analysis is the lack of clean, labeled in-control (IC) data. Directly applying a VAE to a contaminated historical dataset leads to high false-negative rates, as the model inadvertently learns to represent process shifts as part of the nominal distribution. This failure mode undermines the core objective of Phase~I monitoring: establishing a robust baseline for future process behavior. To overcome these limitations, the proposed methodology shifts the focus from simple density estimation to a robust, iterative learning process. By explicitly decoupling the learning of process variability from the influence of anomalies, we can ensure that the resulting reconstruction scores and latent embeddings remain sensitive to the specific deviations that characterize industrial process faults.

To adapt VAEs for anomaly detection and retrospective SPC analysis, the methodology must be reframed so that (i) anomalous observations can be identified and isolated rather than absorbed into the learned manifold, (ii) the reconstruction quality reflects true deviations from IC variation, and (iii) the latent distribution remains stable, Gaussian-like, and representative of the underlying IC process. Achieving these goals requires mechanisms that iteratively refine the training dataset, reduce the influence of anomalous samples, and update the VAE parameters so that the latent space is progressively purified. In this way, both reconstruction errors and latent embeddings become reliable indicators of OC behavior, forming the foundation for robust online monitoring in subsequent phases.

The proposed VSCOUT framework operationalizes this strategy through an integrated pipeline combining ensemble-based outlier detection, VAE training and continued refinement, and latent-space monitoring. The components of this methodology are described in detail in the sections that follow.

\subsection{Step 1: Initial ARD-VAE Training (Relevant Latent Dimensions Extraction)}

\begin{figure}[ht]
  \centering
  \includegraphics[width=0.8\linewidth]{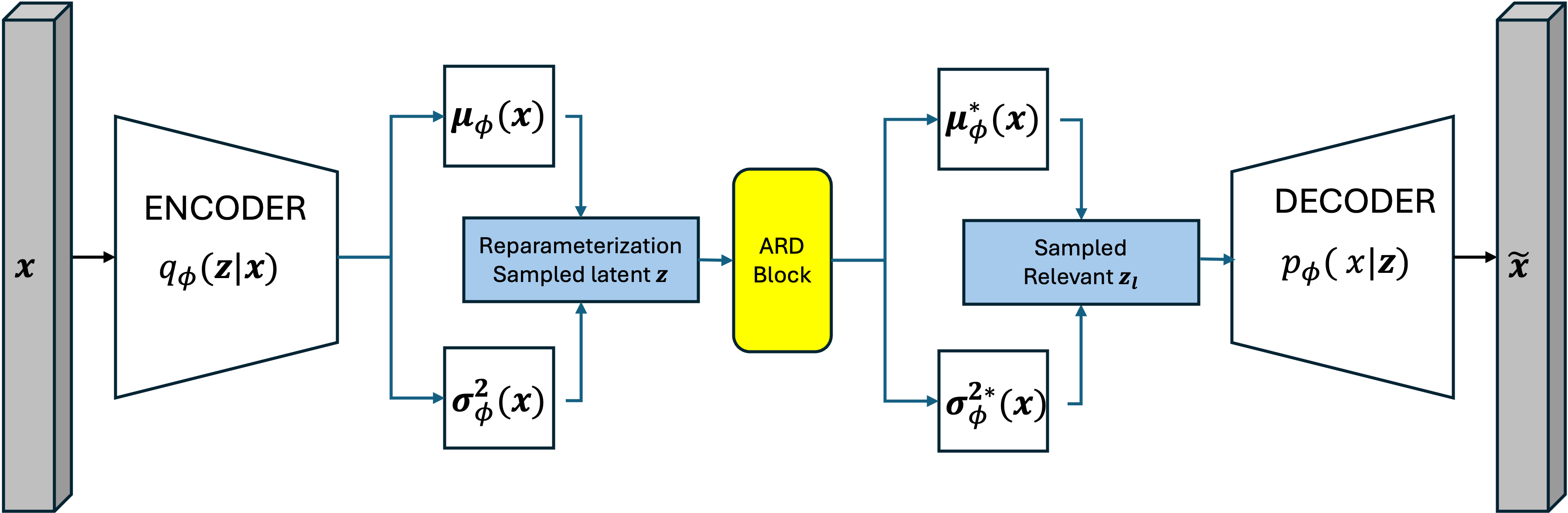}
  \caption{Automatic Relevance Determination VAE (ARD-VAE) architecture.}
  \label{fig:ARDvariationalautoencoder}
\end{figure}

VSCOUT builds upon the VAE framework, but selecting the appropriate size of the latent vector $\mathbf{z} \in \mathbb{R}^d$ is both challenging and essential for accurate SPC monitoring. An overly large latent space can absorb noise and anomalous variation, while an insufficient one may fail to capture intrinsic IC structure. To address this, VSCOUT incorporates the Automatic Relevance Determination VAE (ARD-VAE) framework~\citep{saha2025ard}, which transforms latent-dimension selection from a fixed hyperparameter into a data-driven inference problem (see Figure~\ref{fig:ARDvariationalautoencoder}).

In a standard VAE, the prior over the latent vector is isotropic,
\[
p(\mathbf{z}) = \mathcal{N}(\mathbf{0}, \mathbf{I}),
\]
which implicitly treats all $d$ latent axes as equally important. ARD-VAE replaces this prior with a hierarchical relevance prior that assigns each latent coordinate $z_\ell$ its own precision parameter $\alpha_\ell$:
\[
p(\mathbf{z} \mid \boldsymbol{\alpha})
=
\prod_{\ell=1}^{d}
\mathcal{N}(z_\ell; 0, \alpha_\ell^{-1}),
\qquad
p(\alpha_\ell)
=
\mathrm{Gamma}(a_0, b_0).
\]
Large values of $\alpha_\ell$ force the prior variance $\alpha_\ell^{-1}$ to shrink toward zero, effectively pruning irrelevant latent axes. Integrating out $\alpha_\ell$ yields a heavy-tailed Student-$t$ prior, which naturally encourages sparsity while allowing informative dimensions to remain active.

As in a standard VAE, the encoder outputs a variational posterior
\[
q_\phi(\mathbf{z} \mid \mathbf{x})
=
\mathcal{N}\!\big(
\boldsymbol{\mu}_\phi(\mathbf{x}),
\operatorname{diag}\big(\boldsymbol{\sigma}^2_\phi(\mathbf{x})\big)
\big),
\]
where $\boldsymbol{\mu}_\phi(\mathbf{x})$ and $\boldsymbol{\sigma}^2_\phi(\mathbf{x})$ denote the encoder-estimated mean and variance for a given observation $\mathbf{x}$. Under the ARD prior, however, the KL term in the ELBO becomes dimension-specific and pushes the posterior for irrelevant axes toward zero variance, aligning the posterior with the shrinking prior.

To determine which latent axes are truly relevant, ARD-VAE evaluates both (i) the axis-specific prior precision $\alpha_\ell$ and (ii) the decoder sensitivity to perturbations in the latent means. Decoder sensitivity is captured by the Jacobian
\[
J_\theta(\mathbf{x})
=
\frac{\partial f_\theta(\boldsymbol{\mu}_\phi(\mathbf{x}))}
{\partial \boldsymbol{\mu}_\phi(\mathbf{x})},
\]
whose $\ell$-th column measures how strongly changes in latent dimension $\ell$ affect the reconstruction $\tilde{\mathbf{x}}$. Latent dimensions with small posterior variance, large precision $\alpha_\ell$, and near-zero Jacobian norms are deemed irrelevant.

This yields a principled pruning rule that produces a reduced, relevant latent representation:
\[
\boldsymbol{\mu}^{\ast}_\phi(\mathbf{x})
=
(\mu_{\phi,\ell}(\mathbf{x}) : \ell \in \mathcal{R}),
\qquad
\boldsymbol{\sigma}^{2\ast}_\phi(\mathbf{x})
=
(\sigma^2_{\phi,\ell}(\mathbf{x}) : \ell \in \mathcal{R}),
\]
where $\mathcal{R}$ denotes the index set of relevant latent dimensions identified by the ARD mechanism. The effective latent dimensionality $d_{\text{eff}} = |\mathcal{R}|$ is therefore determined directly from the data rather than specified a priori.

For VSCOUT, this ARD-based latent dimension selection ensures that the learned latent space captures only the stable IC variability while suppressing noisy or anomalous directions. This leads to a compact, interpretable latent representation that forms a reliable basis for both reconstruction-based and latent-distance anomaly scoring in Phase~I and Phase~II SPC monitoring.

\begin{figure}[ht]
  \centering
  \includegraphics[width=0.8\linewidth]{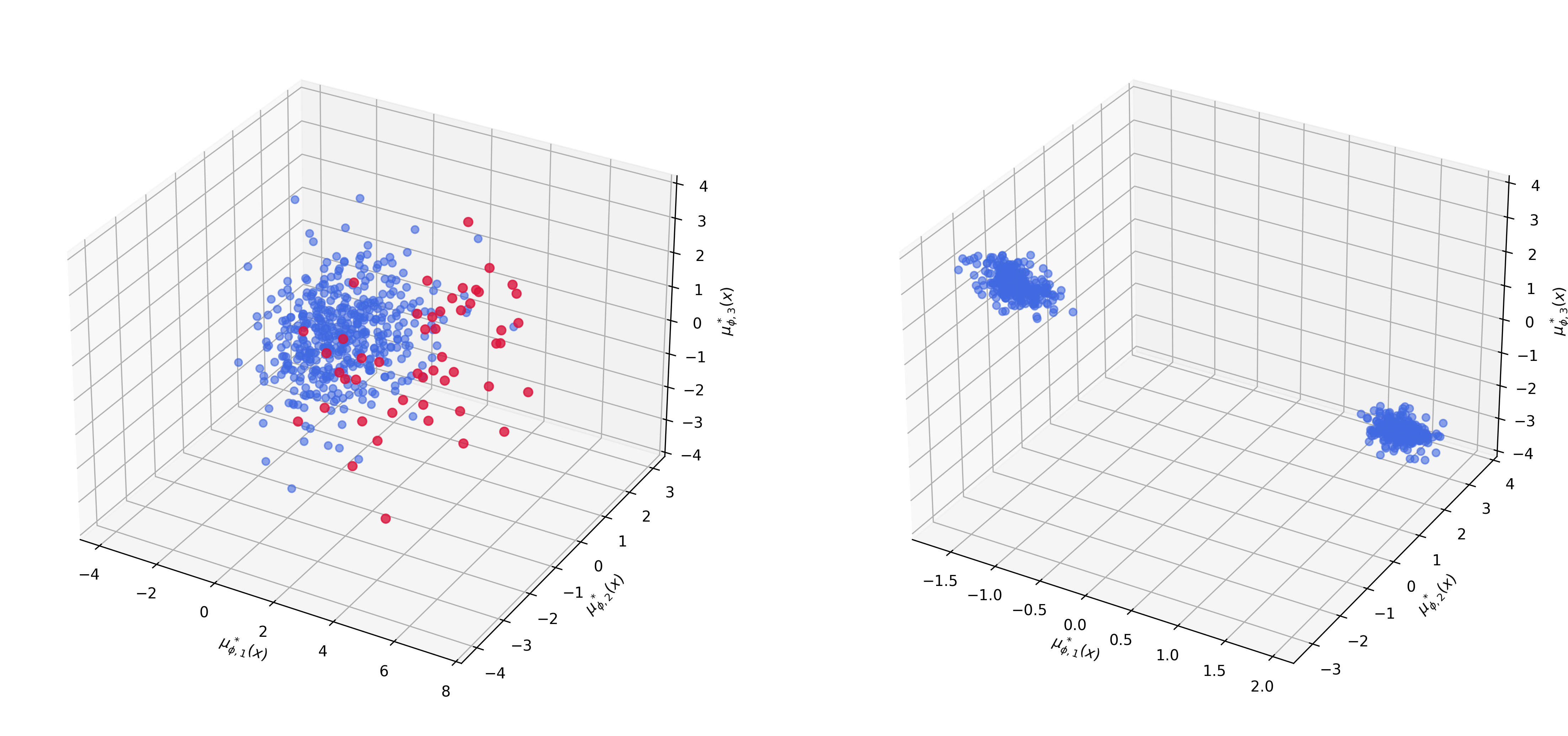}
  \caption{Left: ARD-VAE latent space for a contaminated normal dataset with
  450 in-control observations and 50 shifted observations ($\delta = 1$),
  $p=250$. Right: ARD-VAE latent space for a multimodal dataset with two
  Gaussian clusters (250 observations each) centered at $-5$ and $5$,
  $p=250$.}
  \label{fig:latentdim}
\end{figure}

Figure~\ref{fig:latentdim} provides an empirical illustration of the ARD-VAE
mechanism using a feed-forward architecture with 64 neurons in both the encoder
and decoder and an initial latent dimension of 32. In both examples, the ARD
prior prunes the latent representation to three relevant dimensions
($d_{\mathrm{eff}} = 3$), demonstrating the model’s ability to recover a compact
and interpretable latent space. The left panel corresponds to a contaminated
normal dataset, where the reduced latent space cleanly isolates the shifted
observations while maintaining a coherent representation of the in-control
distribution. The right panel shows a multimodal dataset in which the ARD-VAE
preserves the inherent two-cluster structure despite reducing the dimensionality.

\begin{figure}[ht]
  \centering
  \includegraphics[width=0.7\linewidth]{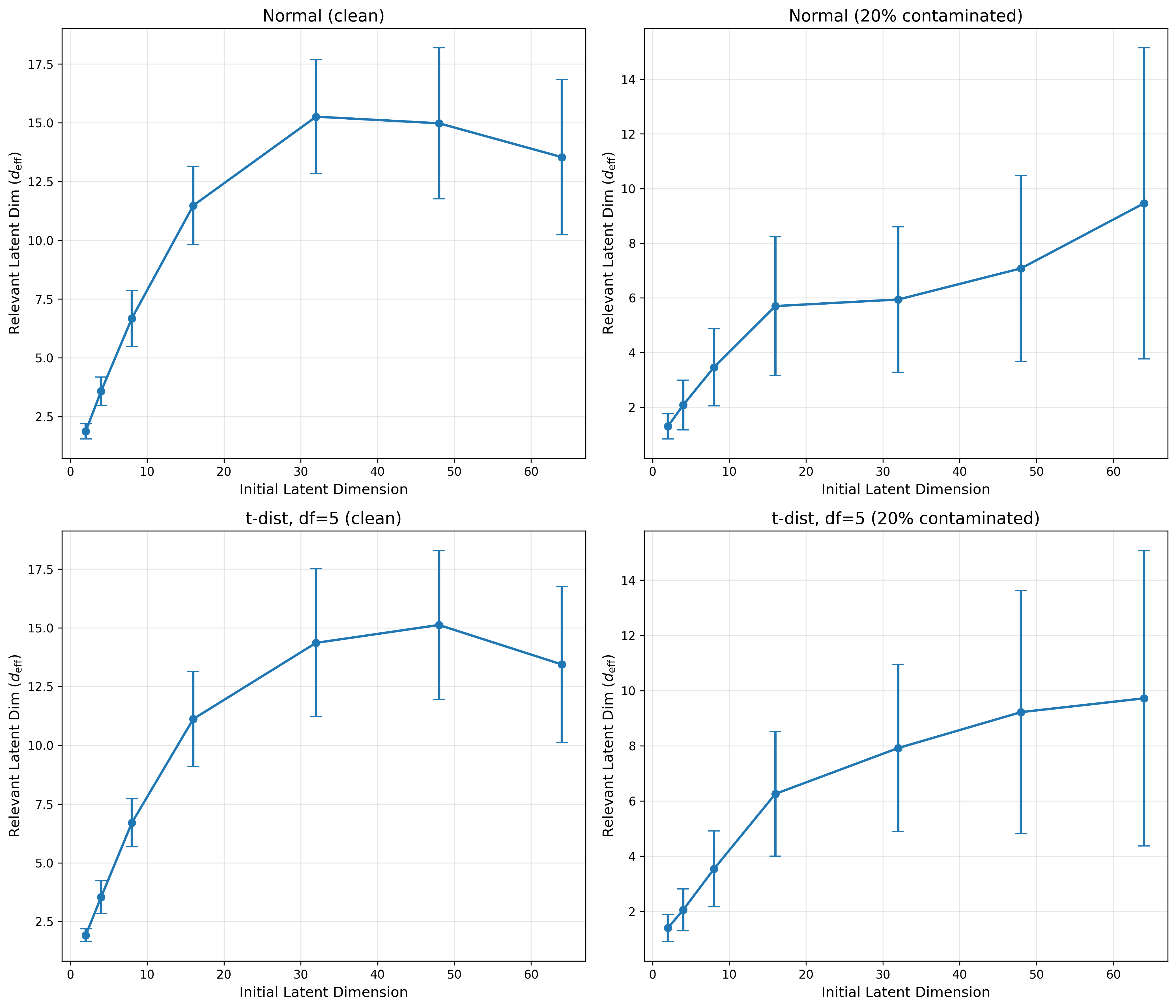}
  \caption{
  Mean effective latent dimensionality $d_{\mathrm{eff}}$ selected by ARD-VAE
  for initial latent sizes ranging from 2 to 64, with error bars showing one
  standard deviation across repeated fits. Top row: clean normal data (left)
  and normal data with 20\% contamination (right). Bottom row: clean
  $t$-distributed data with $\mathrm{df}=5$ (left) and the same distribution
  with 20\% contamination (right).}
  \label{fig:latentdimchange}
\end{figure}

To examine the pruning behavior of the ARD-VAE, we vary the initial latent
dimension from $2$ to $64$ and measure the resulting effective dimensionality
$d_{\mathrm{eff}}$ after training. Figure~\ref{fig:latentdimchange} summarizes
these results across four data regimes: clean normal data, normal data with
20\% contamination, clean $t$-distributed data with $\mathrm{df}=5$, and
$t$-distributed data with 20\% contamination. In all cases, ARD-VAE
substantially reduces the dimensionality of the latent space, even when the
nominal latent capacity is large. For clean datasets, $d_{\mathrm{eff}}$
initially increases and then stabilizes around a value between 12 and 15,
reflecting recovery of the intrinsic structure. Under contamination or heavy
tails, $d_{\mathrm{eff}}$ is consistently smaller, indicating that ARD-VAE
appropriately downweights unstable directions and avoids overfitting to
irregular observations. Across all settings, the model exhibits strong
self-regularization: redundant latent coordinates are suppressed and only
informative axes are retained, demonstrating robustness to distributional shifts
and outlier presence.

\paragraph{Notation Remark.}
For each observation $\mathbf{x}_i$, the ARD-VAE encoder outputs the reduced
latent representation
\[
\mathbf{z}_i
=
\boldsymbol{\mu}^{\ast}_\phi(\mathbf{x}_i)
\in \mathbb{R}^{d_{\mathrm{eff}}},
\]
where $d_{\mathrm{eff}} = |\mathcal{R}|$ is the number of relevant latent
dimensions retained by the ARD mechanism.  
For the entire dataset $X=\{\mathbf{x}_1,\ldots,\mathbf{x}_n\}$, we define the
stacked latent matrix
$\mathbf{Z}
=
\boldsymbol{\mu}^{\ast}_\phi(X)
=
\begin{bmatrix}
\mathbf{z}_1^\top, \ldots,  
\mathbf{z}_n^\top
\end{bmatrix}
\in \mathbb{R}^{n\times d_{\mathrm{eff}}}$. This matrix is the latent representation used in Step~2 for ensemble filtering
and changepoint detection. After refinement in Step~3, the updated latent
vectors are denoted
\[
\mathbf{z}_i^{\mathrm{in}}
=
\boldsymbol{\mu}^{\ast}_\phi(\mathbf{x}_i),
\qquad
\mathbf{x}_i \in X_{\mathrm{in}},
\]
where $X_{\mathrm{in}}$ is the set of observations retained after provisional
ensemble and changepoint filtering.

\subsection{Step 2a: Ensemble-Based Outlier Filter in the Relevant Latent Subspace}

This stage of VSCOUT performs pointwise outlier filtering directly in the
relevant latent subspace. For each observation $\mathbf{x}_i$, the ARD-VAE
encoder outputs the reduced latent representation
\[
\mathbf{z}_i
=
\boldsymbol{\mu}^{\ast}_\phi(\mathbf{x}_i)
\in \mathbb{R}^{d_{\mathrm{eff}}},
\]
where $d_{\mathrm{eff}} = |\mathcal{R}|$ is the number of latent dimensions
retained by the ARD mechanism. Operating in this compact latent space improves
the stability of classical anomaly detectors that may perform poorly in the
original high-dimensional feature space, while also preventing irrelevant
variability from influencing the filtering step.

For the full dataset $X=\{\mathbf{x}_1,\ldots,\mathbf{x}_n\}$, we denote the
stacked latent matrix by
\[
\mathbf{Z}
=
\boldsymbol{\mu}^{\ast}_\phi(X)
=
\begin{bmatrix}
\mathbf{z}_1^\top \\
\vdots \\
\mathbf{z}_n^\top
\end{bmatrix}
\in \mathbb{R}^{n \times d_{\mathrm{eff}}}.
\]
Each detector $D_j$ in the ensemble maps latent vectors to binary decisions,
\[
D_j: \mathbb{R}^{d_{\mathrm{eff}}} \rightarrow \{0,1\},
\]
and VSCOUT accommodates a broad range of detector families, including
distance-based ($k$-Nearest Neighbors~\citep{angiulli2002fast}), 
density-based (Local Outlier Factor~\citep{breunig2000lof}), 
isolation-based (Isolation Forest~\citep{liu2008isolation}), 
distribution-free (ECOD~\citep{li2022ecod}),
histogram-based (HBOS~\citep{goldstein2012histogram}), 
and univariate rules such as boxplot fences when $d_{\mathrm{eff}}=1$.  
Because these detectors operate independently, the ensemble is fully modular
and may be configured to emphasize sensitivity, specificity, or balanced
robustness.

For each observation, the detector outputs are aggregated into a single
provisional ensemble flag
\[
e_i^{(0)} 
=
\mathcal{C}\!\left(
D_1(\mathbf{z}_i), \ldots, D_m(\mathbf{z}_i)
\right),
\qquad e_i^{(0)} \in \{0,1\},
\]
where the consensus function $\mathcal{C}$ may be:
\begin{itemize}
\item \emph{majority voting}: the majority of detectors must flag the observation;
\item \emph{any} (union) rule: at least one detector flags the observation;
\item \emph{all} (intersection) rule: every detector must flag the observation.
\end{itemize}
The ``any'' rule is the most sensitive, whereas the ``all'' rule is the most
conservative; majority voting provides an intermediate balance.

To prevent overly aggressive pruning, VSCOUT allows the user to specify a
maximum contamination level $\alpha$ representing the desired proportion of
flagged observations. The detector-specific thresholds are internally adjusted
so that at most $\alpha n$ points receive $e_i^{(0)} = 1$ in any given iteration.
This constraint stabilizes early refinement rounds and ensures that latent-space
estimation remains anchored to a sufficiently large in-control subset.

Observations with $e_i^{(0)} = 1$ are provisionally removed, and the ARD-VAE
is subsequently retrained on the remaining data. This iterative ensemble
filtering process progressively purifies the latent space, producing a more
accurate and reliable baseline for final anomaly scoring and Phase~I SPC
estimation.


\subsection{Step 2b: Changepoint Detection Filter in the Relevant Latent Subspace}

Beyond pointwise outlier filtering, VSCOUT employs changepoint detection to
identify persistent or structural shifts in the latent representation that may
indicate sustained process deviations. For each observation $\mathbf{x}_i$, the
ARD-VAE encoder provides the reduced latent vector
\[
\mathbf{z}_i = \boldsymbol{\mu}^{\ast}_\phi(\mathbf{x}_i)
\in \mathbb{R}^{d_{\mathrm{eff}}},
\]
and changepoint analysis is performed on the latent magnitude series
\[
s_i = \|\mathbf{z}_i\|_2,
\qquad i = 1,\ldots,n.
\]
While ensemble detectors in Step~2a identify localized or isolated outliers,
structural changes in the distribution of the sequence $\{s_i\}$ reveal
persistent shifts that classical pointwise SPC tools may overlook.

To detect such distributional changes, VSCOUT applies the Pruned Exact Linear
Time (PELT) algorithm~\citep{killick2012optimal} to $\{s_i\}$. Given a segment
cost function $\mathcal{C}(\cdot)$ (commonly an $L_2$ or RBF cost) and a penalty
parameter $\beta>0$, PELT solves
\[
\min_{K,\,\tau_{1:K}}
\left\{
\sum_{k=0}^{K}
\mathcal{C}\!\left(s_{(\tau_k+1):\tau_{k+1}}\right)
\;+\;
\beta K
\right\},
\]
where $0=\tau_0 < \tau_1 < \cdots < \tau_K < \tau_{K+1}=n$ denote changepoint
locations. The penalty $\beta$ controls the effective false-alarm rate: larger
values lead to fewer detected changepoints (more conservative), whereas smaller
values increase sensitivity. In analogy with classical SPC practice, $\beta$ may
be associated with a desired changepoint contamination level
$\alpha_{\mathrm{cp}}$, allowing practitioners to calibrate the conservativeness
of the filter. Implementation is performed using the \texttt{ruptures}
package~\citep{truong2018ruptures}.

From the detected changepoint set $\{\tau_1,\ldots,\tau_K\}$, VSCOUT uses the
earliest changepoint
\[
\tau^\ast = \min\{\tau_k\}
\]
as the onset of a potential sustained shift. A provisional changepoint flag is
assigned by
\[
c_i^{(0)} =
\begin{cases}
1, & i > \tau^\ast, \\
0, & i \le \tau^\ast.
\end{cases}
\]
Observations with $c_i^{(0)} = 1$ are provisionally removed prior to the second
ARD-VAE training stage. This follows the standard retrospective SPC objective of
isolating an in-control segment at the beginning of the dataset before
estimating baseline parameters.

The behavior of PELT is governed primarily by its penalty parameter $\beta$ in
the objective
\[
\sum_{k=0}^{K}
\mathcal{C}\!\left(s_{\tau_k+1:\tau_{k+1}}\right) + \beta K,
\]
where larger $\beta$ discourages additional changepoints and reduces spurious
detections. Although PELT does not use a formal statistical significance level,
$\beta$ acts as an implicit false-alarm controller. Additional stability comes
from the chosen cost function $\mathcal{C}$ (e.g., $L_2$, Gaussian, or RBF) and
a minimum-segment-length constraint, which prevents detections caused by short
oscillations or noise.

The combination of ensemble filtering (Step~2a) and PELT-based changepoint
detection (Step~2b) gives VSCOUT sensitivity to both transient irregularities
and persistent structural deviations. Localized anomalies that violate the
latent manifold are identified by the ensemble, while gradual or sustained
distributional changes in $\{s_i\}$ are captured by PELT. This dual mechanism
produces a robust Phase~I in-control baseline for subsequent SPC monitoring and
enhances detection reliability under complex, high-dimensional process
conditions.

\subsection{Step 3: ARD-VAE Refinement (Second Training Stage)}

After provisional anomalies have been removed through the ensemble and
changepoint filters, VSCOUT performs a second ARD-VAE training step using only
the retained in-control observations. This refinement stage stabilizes the
latent representation by ensuring that the encoder--decoder pair is optimized
exclusively on data believed to originate from the in-control distribution,
thereby reducing latent distortion and improving both reconstruction quality and
probabilistic encoding.

Let $m_i$ denote the provisional inlier mask from Step~2, and define
\[
X_{\mathrm{in}}
=
\{\mathbf{x}_i : m_i = 1\}
\]
as the subset of observations retained for refinement. Rather than
reinitializing the ARD-VAE from scratch, the second-stage optimization
\emph{warm-starts} from the encoder and decoder parameters $(\theta,\phi)$
learned in Step~1. Because these parameters already reflect the global structure
of the full dataset, the refinement stage requires only incremental updates to
remove residual influence from anomalous samples. The parameters are updated
according to
\[
(\theta, \phi)
\leftarrow
(\theta, \phi)
-
\eta \,\nabla_{(\theta,\phi)}\mathcal{L}_{\mathrm{ELBO}},
\]
with learning rate $\eta$ and mini-batch updates, using the same ELBO objective
and reparameterized sampling mechanism as in the initial training.

After refinement, updated latent means
\[
\mathbf{z}_i^{\mathrm{in}}
=
\boldsymbol{\mu}^{\ast}_\phi(\mathbf{x}_i),
\qquad \mathbf{x}_i\in X_{\mathrm{in}},
\]
and corresponding variances
$\boldsymbol{\sigma}_\phi^2(\mathbf{x}_i)$ are computed. New
KL-divergence values are then obtained for each latent coordinate $\ell$:
\[
\mathrm{KL}_\ell
=
\mathbb{E}_{\mathbf{x}\in X_{\mathrm{in}}}
\left[
D_{\mathrm{KL}}\!\left(
q_\phi(z_\ell\mid \mathbf{x})
\;\|\;
p(z_\ell)
\right)
\right].
\]
The relevant-latent set is updated by
\[
\mathcal{R}
=
\big\{
\ell \;:\; \mathrm{KL}_\ell > \tau
\big\},
\].

If the refined set $\mathcal{R}$ differs from the one obtained in Step~1,
VSCOUT preserves the newly selected coordinates for subsequent modeling. If
$\mathcal{R}$ remains unchanged, the encoder and decoder weights from the first
training stage are restored to avoid unnecessary drift in the latent geometry.

This refinement stage yields a cleaned and stable latent subspace representing
the intrinsic variation of the in-control process after removing anomalous or
shifted observations. All subsequent components of VSCOUT---Hotelling’s $T^2$,
reconstruction-error thresholds, ensemble retraining, and changepoint
recalibration---operate on this refined latent representation.

\subsection{Step 4: Final Outlier Determination via Consensus Rule}

The components of VSCOUT---ARD-VAE latent modeling, ensemble filtering, and
changepoint detection---capture complementary aspects of abnormal process
behavior. To produce a stable in-control (IC) reference and a final set of
out-of-control labels, these components are combined through an initial
filtering step followed by a refinement and consensus stage.

After the initial ARD-VAE fit (Step~1), provisional anomalies are flagged by
(i) the ensemble acting on the relevant latent subspace and (ii) changepoint
detection applied to latent magnitude trajectories. For each observation
$\mathbf{x}_i$, the ARD-VAE encoder outputs a reduced latent vector
\[
\mathbf{z}_i
=
\boldsymbol{\mu}^{\ast}_\phi(\mathbf{x}_i)
\in \mathbb{R}^{d_{\mathrm{eff}}},
\qquad
s_i = \|\mathbf{z}_i\|_2,
\]
where $s_i$ denotes the $L_2$ magnitude of the relevant latent representation.
Applying PELT to the sequence $\{s_i\}_{i=1}^n$ yields changepoint locations
$\{\tau_1,\ldots,\tau_K\}$, and the earliest changepoint
\[
\tau^\ast = \min\{\tau_k\}
\]
induces provisional changepoint flags
\[
c_i^{(0)} =
\begin{cases}
1, & i > \tau^\ast, \\
0, & i \le \tau^\ast.
\end{cases}
\]
Simultaneously, the ensemble detectors applied to $\mathbf{z}_i$ produce
provisional ensemble flags $e_i^{(0)}\in\{0,1\}$. These are combined into a
provisional inlier mask
\[
m_i = 1 - \max\{e_i^{(0)},\, c_i^{(0)}\},
\qquad m_i \in \{0,1\},
\]
and the corresponding inlier subset
\[
X_{\mathrm{in}} = \{\mathbf{x}_i : m_i = 1\}
\]
is used for refinement.

In Step~3, the ARD-VAE is retrained on $X_{\mathrm{in}}$. Crucially, the encoder
and decoder parameters are \emph{not} reinitialized: the second-stage
optimization warm-starts from the weights learned during the initial fit.
Thus, gradient descent continues from a parameter state already adapted to the
global structure of the data, and refinement focuses on removing the residual
influence of anomalous samples rather than relearning the entire generative
model. The retrained encoder yields refined latent vectors
\[
\mathbf{z}_i^{\mathrm{in}}
=
\boldsymbol{\mu}^{\ast}_\phi(\mathbf{x}_i),
\qquad i=1,\ldots,n,
\]
and an updated relevant latent set $\mathcal{R}$ when necessary.

Using only the refined inliers ($m_i=1$), VSCOUT then estimates IC statistics in
the final relevant latent subspace:
\[
\boldsymbol{\mu}_{\mathrm{IC}}
=
\frac{1}{n_{\mathrm{in}}}
\sum_{i:m_i=1} \mathbf{z}_i^{\mathrm{in}},
\qquad 
\Sigma_{\mathrm{IC}}
=
\operatorname{Cov}\!\big(\mathbf{z}_i^{\mathrm{in}} : m_i=1\big),
\]
where $n_{\mathrm{in}} = \sum_i m_i$. These define the Hotelling monitoring
statistic
\[
T_i^2
= 
\left(\mathbf{z}_i^{\mathrm{in}} - \boldsymbol{\mu}_{\mathrm{IC}}\right)^\top
\Sigma_{\mathrm{IC}}^{-1}
\left(\mathbf{z}_i^{\mathrm{in}} - \boldsymbol{\mu}_{\mathrm{IC}}\right),
\]
with threshold $h$ set to an empirical or theoretical $(1-\alpha_{T^2})$
quantile under IC conditions.

A brief comment is warranted regarding the stability of the covariance estimate 
$\Sigma_{\mathrm{IC}}^{-1}$ used in $T^{2}$. Although ARD-VAE typically reduces 
the relevant latent dimensionality to a compact $d_{\mathrm{eff}}$, situations 
with limited in-control sample size may still lead to mildly ill-conditioned 
covariance matrices. In all simulation scenarios, the ARD-driven reduction was 
sufficient to ensure numerical stability without explicit shrinkage; however, 
in applications with extremely small $n_{\mathrm{in}}$, practitioners may 
optionally substitute a regularized estimator (e.g., Ledoit–Wolf).

Reconstruction errors
\[
r(\mathbf{x}_i)
=
\|\mathbf{x}_i-\tilde{\mathbf{x}}_i\|_2^2
\]
are computed for all observations, and an IC cutoff $q_\alpha$ (e.g., the 95th
percentile) is estimated from $\{r(\mathbf{x}_i): m_i=1\}$. The ensemble
detectors are refit on the refined latent vectors $\mathbf{z}_i^{\mathrm{in}}$
for $m_i=1$, and then applied to all $\mathbf{z}_i^{\mathrm{in}}$ to obtain
updated ensemble flags $e_i$.

\paragraph{Final Decision Rule.}
For each observation $\mathbf{x}_i$, VSCOUT produces four binary indicators:
\[
c_i \;(\text{changepoint-based}), \qquad
e_i \;(\text{ensemble}), \qquad
u_i = \mathbb{I}(T_i^2 > h), \qquad
q_i = \mathbb{I}(r(\mathbf{x}_i) > q_\alpha).
\]
An observation is labeled out-of-control if at least two of these indicators
agree:
\[
\hat{y}_i = \mathbb{I}\!\left(c_i + e_i + u_i + q_i \;\ge\; 2\right).
\]

This consensus rule reduces sensitivity to spurious errors from any single
component while preserving the ability to detect both abrupt point anomalies and
sustained distributional shifts. By combining statistical distance,
reconstruction behavior, ensemble agreement, and sequential structure in the
refined latent space, VSCOUT delivers robust and interpretable out-of-control
labels suitable for retrospective Phase~I analysis and for establishing reliable
IC baselines for subsequent online monitoring.

\begin{figure}[ht]
  \centering
  \includegraphics[width=0.95\linewidth]{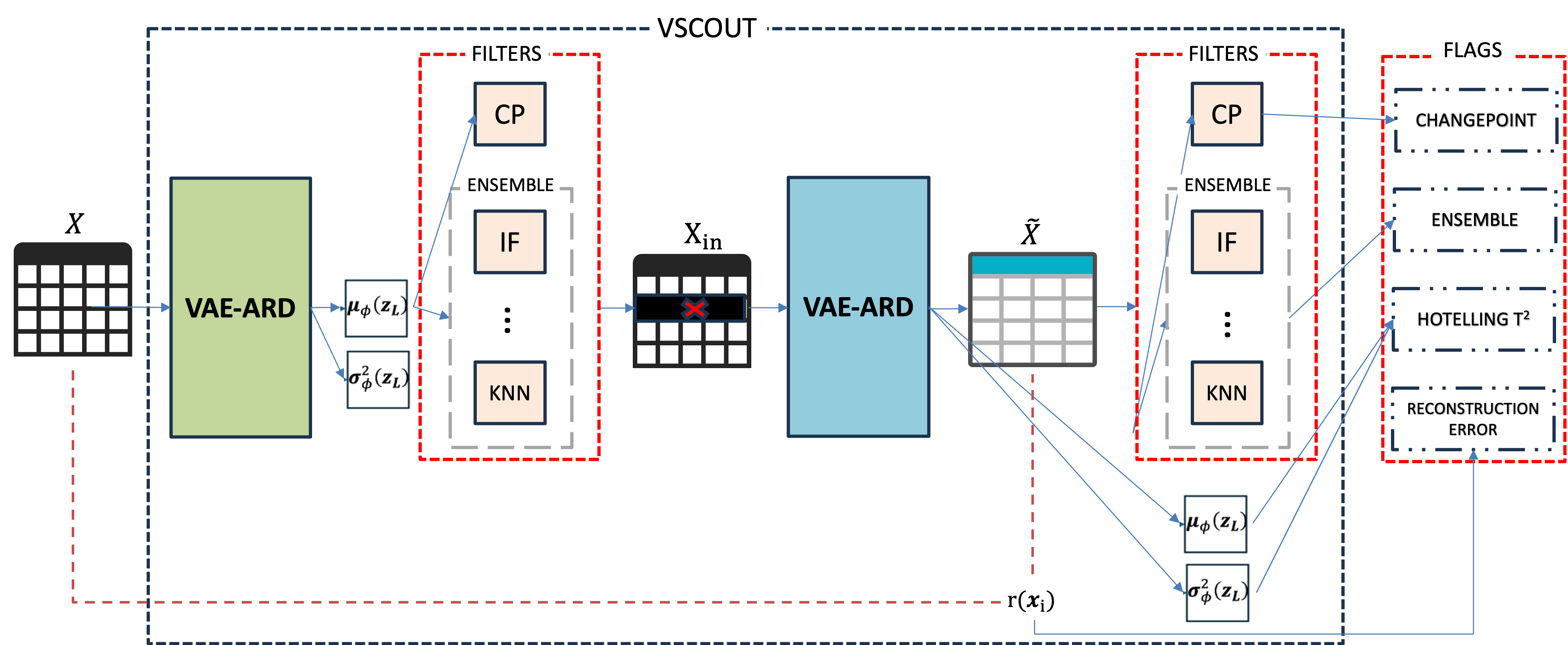}
\caption{Workflow of VSCOUT. }
\label{fig:vscout_pipeline}
\end{figure}

Figure~\ref{fig:vscout_pipeline} provides a high-level overview of the VSCOUT
pipeline, illustrating the sequential interaction between ARD-VAE latent
modeling, ensemble-based filtering, changepoint detection, refinement, and final
consensus scoring. The corresponding pseudocode in
Algorithm~\ref{alg:vscout} formalizes these steps and makes explicit the order
in which latent encoding, provisional anomaly removal, warm-started refinement,
IC parameter estimation, and final indicator evaluation are performed. Steps~1
through~3 implement the iterative purification of the latent space by training
the ARD-VAE, removing provisional anomalies via ensemble and changepoint flags,
and retraining the model on the inlier subset without reinitializing parameters.
Step~4 aggregates four complementary indicators---changepoint flags, ensemble
flags, Hotelling $T^2$ exceedances, and reconstruction-error exceedances---into
a robust 2-of-4 consensus decision rule for the final outlier label $\hat{y}_i$. 

\begin{algorithm}[H]
\caption{VSCOUT: Variational Self-Correcting Outlier Uncovering Technique}
\label{alg:vscout}
\begin{algorithmic}[1]
\REQUIRE Dataset $X \in \mathbb{R}^{n \times p}$; latent size $d$; KL threshold $\lambda$; penalty $\beta$; significance level $\alpha$, architecture neurons.
\ENSURE Outlier labels $\hat{y} \in \{0,1\}^n$.

\STATE \textbf{(Step 1)} Train ARD-VAE on $X$; obtain latent vectors 
$\mathbf{z}_i = \boldsymbol{\mu}^{\ast}_\phi(\mathbf{x}_i)$.

\STATE \textbf{(Step 2b)} Compute magnitudes $s_i=\|\mathbf{z}_i\|_2$; apply PELT with penalty $\beta$; define provisional changepoint flags $c_i^{(0)}$.

\STATE \textbf{(Step 2a)} Apply ensemble detectors to $\mathbf{z}_i$; obtain provisional ensemble flags $e_i^{(0)}$.

\STATE \qquad Form provisional inlier mask $m_i = 1 - \max\{e_i^{(0)}, c_i^{(0)}\}$ and inlier set $X_{\mathrm{in}}=\{\mathbf{x}_i: m_i=1\}$.

\STATE \textbf{(Step 3)} Retrain ARD-VAE on $X_{\mathrm{in}}$; obtain refined latent vectors $\mathbf{z}_i^{\mathrm{in}}$ and update $\mathcal{R}$ if necessary.

\STATE \qquad Estimate IC mean $\boldsymbol{\mu}_{\mathrm{IC}}$ and covariance $\Sigma_{\mathrm{IC}}$ from $\{\mathbf{z}_i^{\mathrm{in}}:m_i=1\}$.

\STATE \qquad Refit ensemble detectors on refined latent vectors.

\STATE \qquad Compute reconstruction errors $r(\mathbf{x}_i)$ and determine cutoff $q_\alpha$.

\FOR{each observation $i=1,\ldots,n$}
  \STATE \quad Compute $c_i = \mathbb{I}(i>\tau^\ast)$, ensemble flag $e_i$, $u_i = \mathbb{I}(T_i^2 > h)$, and
  $q_i = \mathbb{I}(r(\mathbf{x}_i) > q_\alpha)$.
  \STATE \quad Assign final label $\hat{y}_i = \mathbb{I}(c_i + e_i + u_i + q_i \ge 2)$.
\ENDFOR

\RETURN $\hat{y}$
\end{algorithmic}
\end{algorithm}

\subsection{Overall False-Alarm Control}

VSCOUT produces its final out-of-control labels by combining four binary
indicators: changepoint-based flags $c_i$, ensemble outlier votes $e_i$,
latent-space Hotelling exceedances $u_i$ via $T_i^{2}$, and
reconstruction-error exceedances $q_i$. Each component has an associated
in-control false-alarm probability,
\[
\alpha_{\mathrm{cp}}=\Pr(c_i=1),\qquad
\alpha_{\mathrm{ens}}=\Pr(e_i=1),\qquad
\alpha_{T^2}=\Pr(u_i=1),\qquad
\alpha_{\mathrm{rec}}=\Pr(q_i=1).
\]
VSCOUT declares an observation out-of-control when at least two indicators
fire,
\[
y_i=\mathbb{I}(c_i+e_i+u_i+q_i\ge 2),
\qquad
\alpha_{\mathrm{global}}
=\Pr(y_i=1\mid\text{IC}).
\]

\paragraph{Ensemble-level false alarms.}
The ensemble consists of $m$ base detectors, each operating at a nominal
per-detector false-alarm rate $\alpha_0$. Depending on the aggregation rule,
the ensemble inherits an effective rate $\alpha_{\mathrm{ens}}$. Under
approximate independence:
\[
\alpha_{\mathrm{ens}}^{(\mathrm{any})}\approx m\alpha_0,
\qquad
\alpha_{\mathrm{ens}}^{(\mathrm{all})}= \alpha_0^m,
\]
and for majority voting,
\[
\alpha_{\mathrm{ens}}^{(\mathrm{maj})}
=
\sum_{k=\lceil m/2\rceil}^{m}
\binom{m}{k}
\alpha_0^{k}(1-\alpha_0)^{m-k}.
\]
This effective ensemble rate enters directly into the global 2-of-4 decision
rule.

\paragraph{Approximate global false-alarm probability.}
Because the four indicators are designed to be sensitive to different types of
deviations, their in-control exceedances are rare, but not strictly independent.
In particular, the VAE-derived quantities $u_i$ and $q_i$ tend to be positively
correlated, since observations that lie far from the latent mean often yield
larger reconstruction error. Thus, the following approximation serves as a
first-order calibration heuristic rather than a formal independence claim.

Let $I_i^{(j)}\in\{c_i,e_i,u_i,q_i\}$. Under IC conditions, the global 2-of-4
false alarm is dominated by the probability that any two indicators fire:
\[
\alpha_{\mathrm{global}}
\approx
\sum_{(j,k)}
\Pr\!\left(I_i^{(j)}=1,\,I_i^{(k)}=1\right),
\]
where the sum runs over the six indicator pairs. Replacing the joint
probabilities by marginal products yields
\[
\alpha_{\mathrm{global}}
\approx
\alpha_{\mathrm{cp}}\alpha_{\mathrm{ens}}
+ \alpha_{\mathrm{cp}}\alpha_{T^2}
+ \alpha_{\mathrm{cp}}\alpha_{\mathrm{rec}}
+ \alpha_{\mathrm{ens}}\alpha_{T^2}
+ \alpha_{\mathrm{ens}}\alpha_{\mathrm{rec}}
+ \alpha_{T^2}\alpha_{\mathrm{rec}}.
\]

If the non-ensemble components are tuned to a common baseline level
$\alpha_{\mathrm{base}}$, i.e.,
\[
\alpha_{\mathrm{cp}}\approx \alpha_{T^2}\approx \alpha_{\mathrm{rec}}
\approx \alpha_{\mathrm{base}},
\]
then the expression above reduces to
\[
\alpha_{\mathrm{global}}\approx 6\,\alpha_{\mathrm{base}}^{2},
\qquad\Rightarrow\qquad
\alpha_{\mathrm{base}}=\sqrt{\alpha_{\mathrm{global}}/6}.
\]
This provides a transparent analytic guideline for choosing componentwise
thresholds. Because $u_i$ and $q_i$ are often positively correlated in practice,
the analytic approximation is complemented with empirical IC validation to
ensure that the realized global false-alarm rate matches the desired
$\alpha_{\mathrm{global}}$ under finite-sample conditions.

\section{Retrospective Monitoring Experiments}
\label{sec:experiments}

This section evaluates the performance of VSCOUT in a retrospective
(Phase~I) Statistical Process Control (SPC) setting under simulated shift
scenarios. The objective is to assess VSCOUT’s ability to construct a stable
in-control baseline and identify special-cause deviations in high-dimensional
processes.

\vspace{0.5em}
\noindent\textbf{Default VSCOUT Settings.}
Unless otherwise stated, all experiments use a common set of hyperparameters (default values for VSCOUT).
The ARD-VAE component employs a feed-forward encoder--decoder architecture with a single hidden layer of 64 neurons in both the encoder and decoder and a latent dimension of $L=32$ using ReLU activations. The learning
rate is fixed at $\eta=10^{-4}$, and early stopping is applied with a patience of
10 epochs. A latent coordinate $\ell$ is retained if $\mathrm{KL}_\ell>\tau$,
with threshold $\tau=1.0$. Changepoint detection is performed using the PELT
algorithm with penalty $\beta=40$. Latent-space monitoring is based on
Hotelling’s $T^2$ statistic at nominal level $\alpha=0.05$, while
reconstruction-error monitoring uses the same nominal level with thresholds
defined by the corresponding empirical quantiles of the in-control data.

\vspace{0.5em}
\noindent\textbf{Ensemble Filtering.}
VSCOUT applies ensemble filtering to the reduced latent representation
$\mathbf{z}_i=\boldsymbol{\mu}^{\ast}_\phi(\mathbf{x}_i)$ using a diverse set of
classical detectors representing complementary anomaly-detection paradigms,
including Isolation Forest, Local Outlier Factor, ECOD, $k$-nearest neighbors,
HBOS, KDE, and Hotelling’s $T^2$ statistic applied directly in latent space. When
the effective latent dimension reduces to one ($d_{\mathrm{eff}}=1$), a
boxplot-based univariate rule is additionally employed. All detectors are used
with default hyperparameters, and ensemble aggregation follows an
\textit{``any''} rule, whereby an observation is provisionally flagged if
identified as anomalous by at least one detector. This conservative screening
strategy supports VSCOUT’s first-stage objective of removing potentially
contaminated observations prior to ARD-VAE refinement.

\vspace{0.5em}
\noindent\textbf{Evaluation Metrics.} 
Models were evaluated using outlier Detection Rate (Recall), Precision, False
Positive Rate (FPR), Area under the Receiver Operating Curve (AUROC), and Inlier Retention, the latter included to
reflect Phase~I objectives. These metrics balance sensitivity to true
anomalies against the risk of false alarms, a critical trade-off in both general
anomaly detection and retrospective SPC settings.

\subsection{VSCOUT in Retrospective Monitoring}

We evaluate VSCOUT in a retrospective (Phase~I) Statistical Process Control (SPC) setting, where the primary goal is to identify potential special-cause deviations while maintaining nominal Type~I error control under in-control (IC) conditions. In contrast to aggressive detection paradigms, Phase~I analysis emphasizes stability and robustness of the estimated baseline. VSCOUT is compared against several representative Phase~I methodologies, including a distribution-free Phase~I procedure \citep{qiu2013introduction}, a rank-based Phase~I chart \citep{qiu2018some}, a VAE-based monitoring approach \citep{lee2019process}, the Zamba multivariate changepoint procedure \citep{zamba2006multivariate}, and the changepoint detector of \citet{zou2008change}. Diagonal-covariance robust Phase~I charts are also relevant, though their axis-aligned dependence assumptions differ from VSCOUT’s nonlinear latent-space framework. Deep isolation forest \citep{hariri2019extended} and latent-variable selection via ELBD \citep{dong2021elbd} are likewise pertinent competing approaches.

The simulation study spans four data-generating mechanisms: multivariate normal, $t$-distributed ($df=5$), lognormal, and mixed distributions (50\% normal and 50\% $t$). Dimensionality is varied over $p\in\{150,250\}$ with sample size fixed at $n=500$. Out-of-control behavior is introduced via global mean shifts of magnitude $\delta\in\{1,2,3\}$ applied uniformly across all features, while $\delta=0$ corresponds to outlier-free IC data. We additionally vary the contamination fraction $\gamma$, defined as the proportion of observations replaced by shifted samples, considering both low-contamination regimes ($\gamma\in\{0.01,0.03,0.05\}$) and high-contamination regimes ($\gamma\in\{0.10,0.15,0.20\}$). For each unique configuration of distributional form, dimensionality $p$, shift magnitude $\delta$, and contamination fraction $\gamma$, 50 independent Monte Carlo replications are performed for a total of 9,600 replications.

Table~\ref{tab:fpr_no_outliers_by_dist_dim} reports average false positive rates (FPR) across distributions and dimensions. VSCOUT exhibits stable Type~I error control, with an overall mean FPR of 0.0418, indicating slightly conservative yet near-nominal behavior across all regimes. Classical procedures, including DF-Phase~I and Rank-Phase~I, remain closely aligned with the nominal level (\(\approx 0.049{-}0.050\)), while Isolation Forest reflects its user-specified contamination setting. In contrast, the plain VAE baseline shows the greatest FPR inflation (mean \(=0.0610\)), particularly under lognormal and mixed distributions, underscoring the sensitivity of reconstruction-based thresholds when trained on contaminated IC data. Zamba-CP is the most conservative method overall (mean FPR \(=0.0119\)), whereas RMDP displays substantial variability across distributions.

\begin{table}[H]
\centering
\caption{Average False Positive Rate (FPR) in Outlier-Free Data ($\delta=0$), by distribution and dimensionality.}
\label{tab:fpr_no_outliers_by_dist_dim}
\resizebox{0.85\textwidth}{!}{%
\begin{tabular}{llccccccc}
\toprule
Dist. & $p$ & DF-Phase I & IForest & Rank-Phase I & VAE & Zamba-CP & RMDP & VSCOUT \\
\midrule
Normal      & 150 & 0.0490 & 0.0500 & 0.0500 & 0.0491 & 0.0110 & 0.0001 & 0.0290 \\
            & 250 & 0.0490 & 0.0500 & 0.0500 & 0.0513 & 0.0015 & 0.0000 & 0.0310 \\
t ($\text{df}=5$) & 150 & 0.0489 & 0.0500 & 0.0500 & 0.0549 & 0.0181 & 0.0225 & 0.0400 \\
            & 250 & 0.0488 & 0.0500 & 0.0500 & 0.0579 & 0.0208 & 0.0044 & 0.0420 \\
Lognormal   & 150 & 0.0491 & 0.0500 & 0.0500 & 0.0692 & 0.0125 & 0.2333 & 0.0490 \\
            & 250 & 0.0489 & 0.0500 & 0.0500 & 0.0700 & 0.0133 & 0.1042 & 0.0530 \\
Mixed       & 150 & 0.0491 & 0.0500 & 0.0500 & 0.0665 & 0.0160 & 0.0088 & 0.0440 \\
            & 250 & 0.0489 & 0.0500 & 0.0500 & 0.0733 & 0.0021 & 0.0020 & 0.0460 \\
Multimodal  & 150 & 0.0490 & 0.0500 & 0.0500 & 0.0399 & 0.0000 & 0.0000 & 0.0289 \\
            & 250 & 0.0490 & 0.0500 & 0.0500 & 0.0402 & 0.0000 & 0.0000 & 0.0312 \\
\midrule
Average     & --- & 0.0490 & 0.0500 & 0.0500 & 0.0610 & 0.0119 & 0.0469 & 0.0418 \\
\bottomrule
\end{tabular}%
}
\end{table}

\subsubsection{Performance Under Transient Process Shifts ($\delta > 0$)}

Having established that VSCOUT maintains disciplined false-alarm control under
outlier-free conditions ($\gamma=0$), we next examine its behavior in the
presence of transient process shifts. Transient anomalies represent a
particularly challenging Phase~I setting: the shift signal is both weak and
short-lived, and excessive sensitivity may lead to spurious alarms that distort
baseline estimation. Effective retrospective monitoring therefore requires a
method to aggregate intermittent deviations into coherent evidence while
suppressing noise-driven excursions.

Figure~\ref{fig:vscout_transient_example} illustrates VSCOUT’s response to a
representative transient shift scenario generated from an uncorrelated normal
distribution with $n=500$ observations and $p=150$ variables. In this example,
50 observations are randomly contaminated by a brief shift from $N(0,I)$ to
$N(1.5,I)$. Despite the limited magnitude and duration of the shift, VSCOUT
successfully identifies the contaminated segments while remaining stable during
in-control intervals. This behavior highlights VSCOUT’s ability to integrate
weak, intermittent anomaly signals across time and dimensions into consistent
control-chart behavior. Quantitatively, this scenario yields a precision of
0.8596, recall of 0.9800, F1 score of 0.9159, and a false positive rate of just
0.0178, demonstrating both high sensitivity and strong restraint.

\begin{figure}[H]
    \centering
    \includegraphics[width=0.8\textwidth]{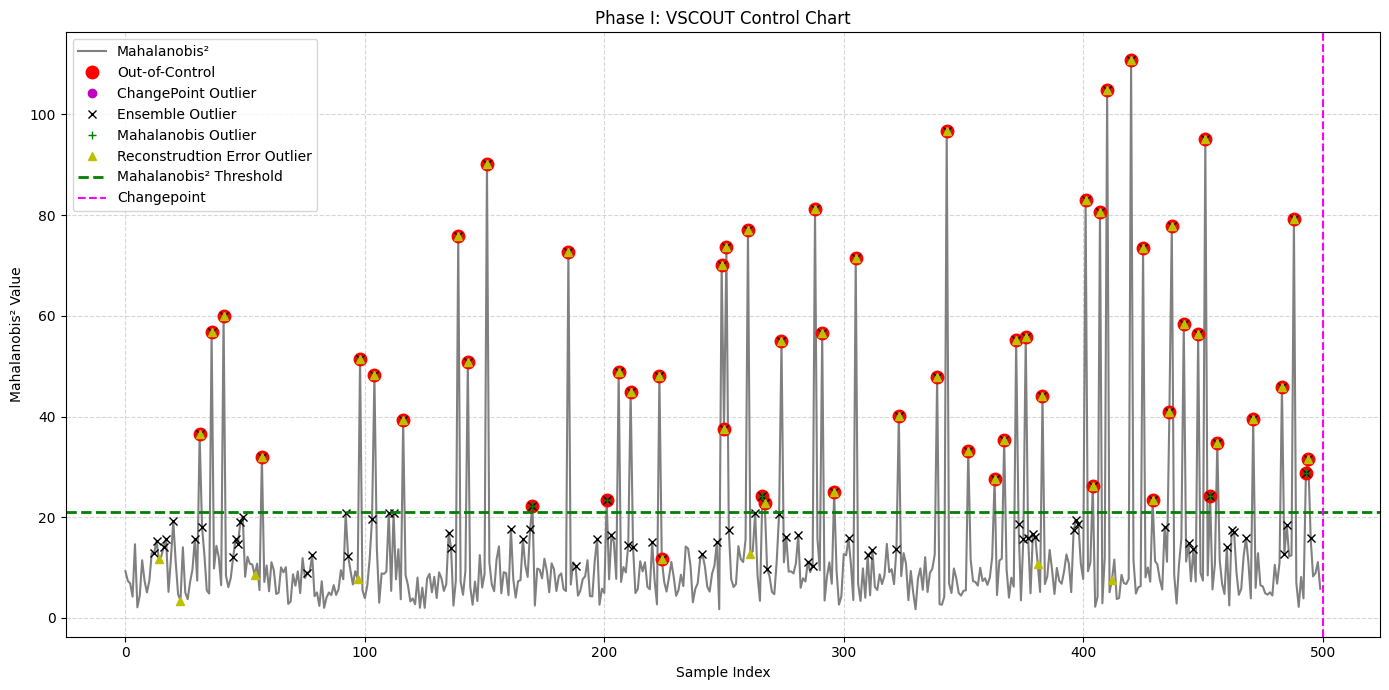}
    \caption{Illustration of VSCOUT control chart applied to data with transient shift from $N(0, I)$ to $N(1.5, I)$ ($\gamma = 0.1$), with $n = 500$, $p = 150$. }
    \label{fig:vscout_transient_example}
\end{figure}

Under low contamination, several methods exhibit strong detection capability,
but important distinctions emerge in terms of stability, robustness, and
computational behavior. VAE achieves the highest average F1 score, driven
primarily by near-perfect recall across most distributions and dimensionalities.
However, this aggressive sensitivity comes at the cost of increased
computational burden and less stable false-positive behavior, particularly in
heavy-tailed and skewed regimes where reconstruction-based thresholds are more
susceptible to distortion. This trade-off is especially apparent in the
LogNormal and Mixed settings, where VAE’s elevated recall is accompanied by
greater variability in FPR and runtime.

In contrast, VSCOUT demonstrates a more balanced detection profile. Across most
distributions, it consistently achieves recall rates exceeding 90\% while
maintaining FPR typically below 5\%, yielding competitive F1 performance without
sacrificing Type~I error control. This behavior reflects the combined effects of
latent-space refinement and ensemble-based screening, which together suppress
spurious variability while preserving sensitivity to genuine transient shifts.
Notably, VSCOUT’s performance remains stable as dimensionality increases from
$p=150$ to $p=250$, indicating robustness to high-dimensional noise and weakly
informative features.

For example, in the Normal ($p=250$) setting, VSCOUT attains an F1 score of 0.6122
with recall of 0.9474, outperforming both Isolation Forest and DF--Phase~I in
terms of the sensitivity--specificity trade-off. Similar patterns are observed
under $t$-distributed and Mixed data, where VSCOUT avoids the sharp degradation
exhibited by classical Phase~I procedures. While Rank--Phase~I also performs
well under low contamination, its effectiveness depends strongly on the validity
of rank-based assumptions and deteriorates under multimodal or heavily skewed
structures.

Classical distribution-free Phase~I methods degrade most noticeably under
non-Gaussian conditions, where heavy tails and skewness violate implicit model
assumptions, leading to near-zero recall in several settings. Zamba--CP remains
highly conservative across all low-contamination scenarios, producing almost no
detections and consequently yielding negligible F1 scores despite excellent
false-alarm control. RMDP shows isolated strengths, particularly under Normal
data, but suffers from substantial variability across distributions and
consistently higher runtime, limiting its practical appeal in large-scale Phase~I
studies.

Taken together, these results suggest that while several methods are capable of
detecting transient shifts under mild contamination, VSCOUT offers one of the
most reliable compromises between sensitivity, specificity, and computational
efficiency. Its stable behavior across distributions and dimensions makes it
particularly well suited for retrospective monitoring scenarios in which
transient anomalies must be identified without destabilizing the in-control
baseline.

\begin{table}[H]
\centering
\scriptsize
\caption{Average performance metrics at low contamination levels ($\gamma \in \{0.01,0.03,0.05\}$) for transient mean shifts.}
\label{tab:multi_metric_low_sorted_clean}
\resizebox{0.88\textwidth}{!}{
\begin{tabular}{lllcccccccc}
\toprule
Distribution & Size $(n,p)$ & Metric 
& DF-Phase I & IForest & Rank-Phase I & VAE & Zamba-CP & RMDP & VSCOUT \\
\midrule

Normal & (500, 150) 
& F1        & 0.0000 & 0.4266 & 0.6781 & 0.5651 & 0.0030 & 0.9975 & 0.6035 \\
& & FPR       & 0.0507 & 0.0735 & 0.0220 & 0.0378 & 0.0010 & 0.0000 & 0.0262 \\
& & Precision & 0.0000 & 0.2912 & 0.5771 & 0.4115 & 0.0078 & 1.0000 & 0.4780 \\
& & Recall    & 0.0000 & 0.9970 & 0.9997 & 0.9989 & 0.0033 & 0.9953 & 0.9173 \\
& & Runtime (s) & 1.6057 & 0.0677 & 0.0972 & 1.9419 & 0.8585 & 3.8525 & 6.7294 \\
\midrule
Normal & (500, 250)
& F1        & 0.0000 & 0.4274 & 0.6785 & 0.5568 & 0.0000 & 0.7636 & 0.6122 \\
& & FPR       & 0.0507 & 0.0734 & 0.0220 & 0.0393 & 0.0000 & 0.0000 & 0.0273 \\
& & Precision & 0.0000 & 0.2918 & 0.5774 & 0.4030 & 0.0000 & 0.7822 & 0.4792 \\
& & Recall    & 0.0000 & 0.9988 & 1.0000 & 0.9988 & 0.0000 & 0.7529 & 0.9474 \\
& & Runtime (s) & 2.6697 & 0.0704 & 0.1578 & 2.2095 & 2.1385 & 9.7310 & 8.5476 \\
\midrule

$t$ (df=5) & (500, 150)
& F1 & 0.0001 & 0.4062 & 0.5955 & 0.5306 & 0.0000 & 0.6396 & 0.4562 \\
& & FPR & 0.0506 & 0.0750 & 0.0259 & 0.0429 & 0.0000 & 0.0146 & 0.0364 \\
& & Precision & 0.0001 & 0.2769 & 0.5038 & 0.3812 & 0.0000 & 0.5662 & 0.3504 \\
& & Recall & 0.0004 & 0.9562 & 0.8925 & 0.9832 & 0.0000 & 0.8123 & 0.7632 \\
& & Runtime (s) & 1.5531 & 0.0784 & 0.0971 & 2.5903 & 0.8517 & 3.7634 & 9.8595 \\
\midrule
$t$ (df=5) & (500, 250)
& F1 & 0.0000 & 0.4133 & 0.6418 & 0.5360 & 0.0017 & 0.6194 & 0.5179 \\
& & FPR & 0.0506 & 0.0744 & 0.0238 & 0.0425 & 0.0008 & 0.0025 & 0.0334 \\
& & Precision & 0.0000 & 0.2819 & 0.5433 & 0.3851 & 0.0019 & 0.6977 & 0.3996 \\
& & Recall & 0.0000 & 0.9686 & 0.9585 & 0.9923 & 0.0031 & 0.6291 & 0.8398 \\
& & Runtime (s) & 2.5569 & 0.0908 & 0.1577 & 6.1467 & 2.1502 & 9.7411 & 11.3769 \\
\midrule

LogNormal & (500, 150)
& F1 & 0.3825 & 0.4282 & 0.6755 & 0.7720 & 0.0174 & 0.2152 & 0.4798 \\
& & FPR & 0.0352 & 0.0733 & 0.0222 & 0.0125 & 0.0021 & 0.2086 & 0.0543 \\
& & Precision & 0.3049 & 0.2924 & 0.5745 & 0.6957 & 0.0595 & 0.1242 & 0.3406 \\
& & Recall & 0.6598 & 1.0000 & 0.9970 & 0.9951 & 0.0146 & 1.0000 & 0.9656 \\
& & Runtime (s) & 1.5651 & 0.0640 & 0.0974 & 1.9652 & 0.8605 & 3.6044 & 6.5718 \\
\midrule
LogNormal & (500, 250)
& F1 & 0.3748 & 0.4282 & 0.6783 & 0.7922 & 0.0133 & 0.3416 & 0.5031 \\
& & FPR & 0.0355 & 0.0733 & 0.0220 & 0.0105 & 0.0012 & 0.1052 & 0.0483 \\
& & Precision & 0.2982 & 0.2924 & 0.5772 & 0.7208 & 0.0373 & 0.2145 & 0.3630 \\
& & Recall & 0.6496 & 1.0000 & 0.9998 & 0.9988 & 0.0125 & 1.0000 & 0.9627 \\
& & Runtime (s) & 2.6154 & 0.0677 & 0.1581 & 2.1961 & 2.1299 & 12.5166 & 8.4695 \\
\midrule

Mixed & (500, 150)
& F1 & 0.0000 & 0.3976 & 0.5668 & 0.5034 & 0.0039 & 0.4503 & 0.4040 \\
& & FPR & 0.0506 & 0.0756 & 0.0272 & 0.0493 & 0.0004 & 0.0536 & 0.0531 \\
& & Precision & 0.0000 & 0.2709 & 0.4797 & 0.3545 & 0.0082 & 0.3285 & 0.2905 \\
& & Recall & 0.0000 & 0.9376 & 0.8493 & 0.9897 & 0.0026 & 0.8904 & 0.7928 \\
& & Runtime (s) & 1.5671 & 0.0793 & 0.0971 & 2.5545 & 0.8663 & 3.6775 & 9.8029 \\
\midrule
Mixed & (500, 250)
& F1 & 0.0000 & 0.4068 & 0.5855 & 0.4894 & 0.0002 & 0.5348 & 0.4379 \\
& & FPR & 0.0506 & 0.0749 & 0.0264 & 0.0529 & 0.0011 & 0.0123 & 0.0518 \\
& & Precision & 0.0000 & 0.2773 & 0.4955 & 0.3411 & 0.0007 & 0.5057 & 0.3137 \\
& & Recall & 0.0000 & 0.9591 & 0.8764 & 0.9948 & 0.0001 & 0.6904 & 0.8521 \\
& & Runtime (s) & 2.5728 & 0.0859 & 0.1577 & 3.6048 & 2.1502 & 9.6398 & 13.9198 \\
\midrule

Multimodal & (500, 150)
& F1 & 0.6942 & 0.4282 & 0.0000 & 0.4844 & 0.0000 & 0.0000 & 0.6263 \\
& & FPR & 0.0200 & 0.0733 & 0.0527 & 0.0308 & 0.0000 & 0.0000 & 0.0294 \\
& & Precision & 0.5997 & 0.2924 & 0.0000 & 0.3734 & 0.0000 & 0.0000 & 0.4794 \\
& & Recall & 0.9997 & 1.0000 & 0.0000 & 0.7618 & 0.0000 & 0.0000 & 0.9937 \\
& & Runtime (s) & 1.5710 & 0.0692 & 0.0969 & 1.7580 & 0.8582 & 3.6633 & 6.7035 \\
\midrule
Multimodal & (500, 250)
& F1 & 0.6943 & 0.4282 & 0.0000 & 0.5172 & 0.0000 & 0.0000 & 0.6223 \\
& & FPR & 0.0200 & 0.0733 & 0.0527 & 0.0298 & 0.0000 & 0.0000 & 0.0307 \\
& & Precision & 0.5998 & 0.2924 & 0.0000 & 0.3989 & 0.0000 & 0.0000 & 0.4748 \\
& & Recall & 0.9997 & 1.0000 & 0.0000 & 0.8120 & 0.0000 & 0.0000 & 0.9992 \\
& & Runtime (s) & 2.5746 & 0.0706 & 0.1577 & 1.9427 & 2.1383 & 9.7407 & 7.5452 \\

\bottomrule
\end{tabular}}
\end{table}

To assess robustness more systematically, we consider transient mean shifts
across two contamination regimes: low contamination
($\gamma\in\{0.01,0.03,0.05\}$) and high contamination
($\gamma\in\{0.10,0.15,0.20\}$). Five data-generating mechanisms are examined
(Normal, $t$ with $df=5$, LogNormal, Mixed, and Multimodal), along with two
dimensionalities ($p=150$ and $p=250$). Average performance metrics are reported
in Tables~\ref{tab:multi_metric_low_sorted_clean} and
\ref{tab:multi_metric_high_clean}.

Under low contamination, several methods exhibit strong detection capability,
but important distinctions emerge. VAE achieves the highest average F1 score,
driven largely by near-perfect recall; however, this performance is accompanied
by increased computational cost and less stable false-positive behavior under
non-Gaussian regimes. VSCOUT consistently delivers recall rates exceeding 90\%
across most distributions while maintaining FPR typically below 5\%, resulting
in competitive F1 performance across Normal, LogNormal, and Mixed settings. For
example, in the Normal ($p=250$) case, VSCOUT attains an F1 score of 0.6122 with
recall of 0.9474, outperforming both IForest and DF--Phase~I in terms of the
sensitivity--specificity trade-off. Rank--Phase~I also performs well in this
regime when its rank-based assumptions align with the data structure. In
contrast, DF--Phase~I deteriorates under heavy-tailed and skewed distributions,
while Zamba--CP remains overly conservative, exhibiting near-zero recall across
most settings. RMDP shows isolated strengths but suffers from substantial
variability and consistently higher runtime.

\begin{table}[H]
\centering
\scriptsize
\caption{Average performance metrics at high contamination levels ($\gamma \in \{0.10,0.15,0.20\}$) for transient mean shifts.}
\label{tab:multi_metric_high_clean}
\resizebox{0.88\textwidth}{!}{
\begin{tabular}{lllcccccccc}
\toprule
Distribution & Size $(n,p)$ & Metric 
& DF-Phase I & IForest & Rank-Phase I & VAE & Zamba-CP & RMDP & VSCOUT \\
\midrule

Normal & (500, 150)
& F1        & 0.0000 & 0.7965 & 0.5303 & 0.3225 & 0.0000 & 0.0026 & 0.6507 \\
& & FPR       & 0.0580 & 0.0041 & 0.0000 & 0.0389 & 0.0000 & 0.0000 & 0.0212 \\
& & Precision & 0.0000 & 0.9652 & 1.0000 & 0.4327 & 0.0000 & 0.0511 & 0.7460 \\
& & Recall    & 0.0000 & 0.7022 & 0.3687 & 0.3057 & 0.0000 & 0.0013 & 0.6394 \\
& & Runtime (s) & 1.7906 & 0.0684 & 0.1250 & 2.1077 & 0.9052 & 3.5257 & 7.0242 \\
\midrule
Normal & (500, 250)
& F1        & 0.0000 & 0.8042 & 0.5303 & 0.3242 & 0.0000 & 0.0000 & 0.6880 \\
& & FPR       & 0.0580 & 0.0030 & 0.0000 & 0.0402 & 0.0000 & 0.0000 & 0.0207 \\
& & Precision & 0.0000 & 0.9742 & 1.0000 & 0.4238 & 0.0000 & 0.0000 & 0.7645 \\
& & Recall    & 0.0000 & 0.7091 & 0.3687 & 0.3138 & 0.0000 & 0.0000 & 0.6913 \\
& & Runtime (s) & 2.9463 & 0.0750 & 0.2048 & 2.6477 & 2.1849 & 8.8990 & 8.9833 \\
\midrule

$t$ (df=5) & (500, 150)
& F1 & 0.0001 & 0.7281 & 0.4712 & 0.2752 & 0.0000 & 0.0212 & 0.5613 \\
& & FPR & 0.0580 & 0.0138 & 0.0067 & 0.0405 & 0.0000 & 0.0009 & 0.0304 \\
& & Precision & 0.0001 & 0.8831 & 0.8889 & 0.3897 & 0.0000 & 0.1617 & 0.6421 \\
& & Recall & 0.0001 & 0.6414 & 0.3276 & 0.2485 & 0.0000 & 0.0120 & 0.5542 \\
& & Runtime (s) & 1.7584 & 0.0813 & 0.1253 & 4.6474 & 0.9073 & 3.4729 & 9.5022 \\
\midrule
$t$ (df=5) & (500, 250)
& F1 & 0.0001 & 0.7497 & 0.4998 & 0.2916 & 0.0000 & 0.0014 & 0.6344 \\
& & FPR & 0.0580 & 0.0105 & 0.0035 & 0.0408 & 0.0000 & 0.0001 & 0.0269 \\
& & Precision & 0.0002 & 0.9101 & 0.9420 & 0.4041 & 0.0000 & 0.0344 & 0.7089 \\
& & Recall & 0.0001 & 0.6602 & 0.3476 & 0.2687 & 0.0000 & 0.0007 & 0.6329 \\
& & Runtime (s) & 2.9152 & 0.0883 & 0.2046 & 3.6755 & 2.1968 & 8.9617 & 19.7778 \\
\midrule

LogNormal & (500, 150)
& F1 & 0.1121 & 0.8213 & 0.5302 & 0.6725 & 0.0096 & 0.6311 & 0.8847 \\
& & FPR & 0.0454 & 0.0007 & 0.0000 & 0.0015 & 0.0009 & 0.1948 & 0.0222 \\
& & Precision & 0.2158 & 0.9940 & 0.9999 & 0.9816 & 0.1535 & 0.4669 & 0.8650 \\
& & Recall & 0.0774 & 0.7248 & 0.3686 & 0.5428 & 0.0051 & 1.0000 & 0.9205 \\
& & Runtime (s) & 1.7423 & 0.0637 & 0.1253 & 1.9732 & 0.9126 & 3.4518 & 6.9867 \\
\midrule
LogNormal & (500, 250)
& F1 & 0.1136 & 0.8213 & 0.5303 & 0.6823 & 0.0053 & 0.7340 & 0.8961 \\
& & FPR & 0.0453 & 0.0007 & 0.0000 & 0.0008 & 0.0003 & 0.1211 & 0.0209 \\
& & Precision & 0.2189 & 0.9940 & 1.0000 & 0.9898 & 0.0861 & 0.5830 & 0.8719 \\
& & Recall & 0.0784 & 0.7248 & 0.3687 & 0.5566 & 0.0027 & 1.0000 & 0.9357 \\
& & Runtime (s) & 2.8680 & 0.0712 & 0.2051 & 2.3809 & 2.1842 & 10.3278 & 9.1787 \\
\midrule

Mixed & (500, 150)
& F1 & 0.0000 & 0.7131 & 0.4625 & 0.2850 & 0.0000 & 0.0481 & 0.5710 \\
& & FPR & 0.0580 & 0.0160 & 0.0076 & 0.0452 & 0.0000 & 0.0017 & 0.0404 \\
& & Precision & 0.0000 & 0.8643 & 0.8731 & 0.3783 & 0.0000 & 0.1820 & 0.6148 \\
& & Recall & 0.0000 & 0.6287 & 0.3214 & 0.2709 & 0.0000 & 0.0332 & 0.5803 \\
& & Runtime (s) & 1.7618 & 0.0810 & 0.1255 & 4.5020 & 0.9131 & 3.4244 & 9.5293 \\
\midrule
Mixed & (500, 250)
& F1 & 0.0001 & 0.7214 & 0.4887 & 0.2985 & 0.0000 & 0.0029 & 0.5935 \\
& & FPR & 0.0579 & 0.0147 & 0.0046 & 0.0465 & 0.0000 & 0.0001 & 0.0390 \\
& & Precision & 0.0001 & 0.8754 & 0.9227 & 0.3904 & 0.0000 & 0.0548 & 0.6356 \\
& & Recall & 0.0000 & 0.6353 & 0.3397 & 0.2878 & 0.0000 & 0.0015 & 0.6104 \\
& & Runtime (s) & 2.9691 & 0.0878 & 0.2057 & 3.6811 & 2.1976 & 9.7347 & 19.7218 \\
\midrule

Multimodal & (500, 150)
& F1 & 0.4270 & 0.7229 & 0.0000 & 0.2313 & 0.0000 & 0.0000 & 0.8191 \\
& & FPR & 0.0000 & 0.0145 & 0.0578 & 0.0336 & 0.0000 & 0.0000 & 0.0083 \\
& & Precision & 1.0000 & 0.8680 & 0.0000 & 0.3168 & 0.0000 & 0.0000 & 0.8691 \\
& & Recall & 0.2890 & 0.6406 & 0.0000 & 0.2032 & 0.0000 & 0.0000 & 0.8035 \\
& & Runtime (s) & 1.7628 & 0.0769 & 0.1250 & 2.3330 & 0.9081 & 3.5358 & 7.0006 \\
\midrule
Multimodal & (500, 250)
& F1 & 0.4284 & 0.7290 & 0.0000 & 0.2520 & 0.0000 & 0.0000 & 0.8258 \\
& & FPR & 0.0000 & 0.0138 & 0.0578 & 0.0316 & 0.0000 & 0.0000 & 0.0081 \\
& & Precision & 1.0000 & 0.8805 & 0.0000 & 0.3621 & 0.0000 & 0.0000 & 0.8752 \\
& & Recall & 0.2902 & 0.6416 & 0.0000 & 0.2115 & 0.0000 & 0.0000 & 0.8106 \\
& & Runtime (s) & 2.9104 & 0.0841 & 0.2052 & 2.5939 & 2.1901 & 9.0117 & 8.4727 \\

\bottomrule
F1 Avg & & & 0.1139 & 0.7620 & 0.4043 & 0.3453 & 0.0015 & 0.1441 & 0.7099 \\
F1 Rank & & & (6) & (1) & (3) & (4) & (7) & (5) & (2) \\
\bottomrule
\end{tabular}}
\end{table}

\subsubsection{Performance Under Sustained Process Shifts ($\gamma > 0$)}

To evaluate robustness under persistent and short-term anomalies, we examine model performance under sustained process shifts as well. Figure~\ref{fig:vscout_transient_example} presents an example of VSCOUT applied to a transient shift scenario, where the data temporarily shifts from a $N(0, I)$ baseline to $N(1.5, I)$. The model accurately isolates the contaminated batches with no false alarms outside the shift region. Quantitatively, this example achieves a precision of 0.9434, recall of 1.0000, and F1 score of 0.9709, with a false positive rate of just 0.0257, showcasing VSCOUT’s ability to combine high sensitivity with specificity, even during short-lived anomalies.

\begin{figure}[H]
\centering
\includegraphics[width=0.8\textwidth]{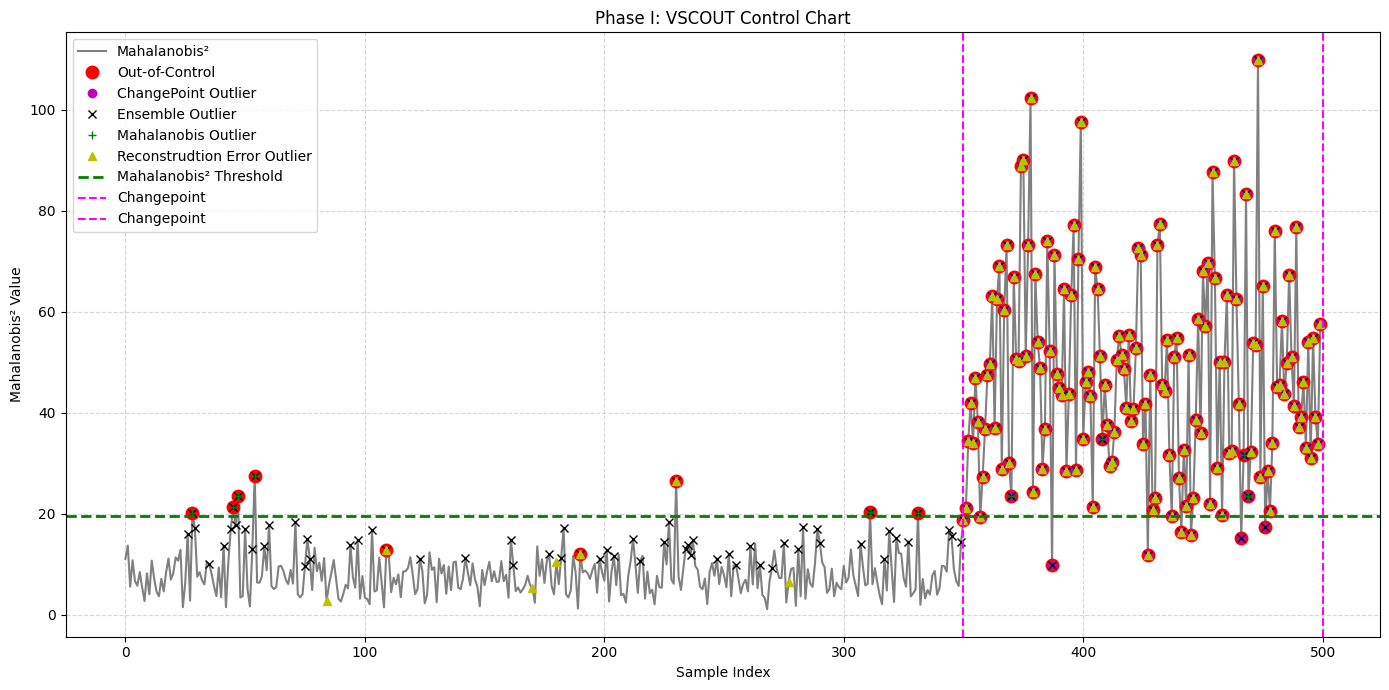}
\caption{Illustration of VSCOUT control chart applied to a transient process shift: $N(0, I) \rightarrow N(1.5, I)$, with $n = 500$, $p = 150$.}
\label{fig:vscout_transient_example}
\end{figure}

We further analyze performance under sustained shifts, reporting average results across both low ($\gamma \in \{0.01,0.03,0.05\}$) and high ($\gamma \in \{0.10,0.15,0.20\}$) contamination levels. Under low contamination (Table~\ref{tab:sustained_multi_metric_low_clean}), VSCOUT ranks among the top-performing methods and maintains consistently strong F1 scores across all distributions and dimensionalities. A key observation is that VSCOUT’s performance remains stable as the dimension increases from $p=150$ to $p=250$, whereas several baseline methods (notably Rank--Phase~I and VAE) show visible degradation with increasing $p$. For instance, in the $t$-distribution at $p=250$, VSCOUT improves its F1 score to 0.5612 compared to 0.5111 at $p=150$, reflecting increased robustness in the presence of heavy tails. Even in the Normal and Mixed settings, where some classical methods perform competitively, VSCOUT maintains F1 values in the $0.54$--$0.66$ range, demonstrating reliable detection even when the sustained shift affects only a small fraction of the data.

VSCOUT is especially effective in non-Gaussian regimes. In the LogNormal distribution---a setting that typically challenges classical distance-based detectors due to skewness and multiplicative noise---VSCOUT achieves F1 scores above 0.86 in both dimensions, with FPR below 2.1\%. In contrast, DF--Phase~I frequently collapses to near-zero recall, Rank--Phase~I exhibits inflated FPR despite high recall, and RMDP suffers from low precision due to overly aggressive flagging. Zamba--CP maintains perfect precision and recall in some Normal settings but fails almost entirely in other distributions, producing near-zero detection under LogNormal or Mixed cases. Across all these regimes, VSCOUT preserves a balanced precision--recall profile, reflecting the benefits of latent refinement, reconstruction diagnostics, and ensemble consensus.

At high contamination levels (Table~\ref{tab:multi_metric_high_sustained_clean}), the detection task becomes significantly more challenging. Several methods deteriorate sharply: Isolation Forest exhibits near-100\% FPR across all distributions, Rank--Phase~I loses power as $\gamma$ increases, and DF--Phase~I remains largely ineffective regardless of dimension or distribution. VAE maintains moderate precision but often fails to achieve sufficient recall, particularly in heavy-tailed or multimodal settings where its reconstruction model becomes unstable. These trends are most pronounced in the $t$~and Mixed distributions at $p=250$, where VAE’s recall drops to 0.2748 and 0.2802, respectively.

Despite these challenges, VSCOUT remains one of the most reliable performers, achieving the second-highest overall F1 average (0.7483) and displaying consistently strong precision--recall balance across all distributions. Its robustness is especially evident in heavy-tailed or skewed settings. For instance, in the LogNormal ($p=250$) case, VSCOUT attains an F1 of 0.8920 with recall above 0.97 and FPR below 3.4\%, outperforming all classical and deep baselines. Even in difficult multimodal scenarios—where many competing methods collapse to near-zero precision or recall—VSCOUT maintains F1 values of 0.5858 ($p=150$) and 0.6834 ($p=250$), reflecting the stabilizing effect of its latent refinement and consensus filtering.

Although Zamba--CP achieves the highest mean F1, its seemingly perfect recall in several settings is largely an artifact of its tendency to flag nearly all points under high contamination, leading to artificially inflated performance metrics that do not translate to practical SPC use. In contrast, VSCOUT preserves controlled FPR along with high precision and recall, providing an interpretable and stable detector even when up to 20\% of observations are contaminated. This reliability under aggressive contamination highlights VSCOUT’s suitability for challenging Phase~I environments where both subtle and dense process shifts must be identified without compromising Type~I error control.

\begin{table}[H]
\centering
\scriptsize
\caption{Average performance metrics at low contamination levels ($\gamma \in \{0.01,0.03,0.05\}$) for sustained mean shifts.}
\label{tab:sustained_multi_metric_low_clean}
\resizebox{0.88\textwidth}{!}{
\begin{tabular}{lllccccccc}
\toprule
Distribution & Size $(n,p)$ & Metric 
& DF-Phase I & IForest & Rank-Phase I & VAE & Zamba-CP & RMDP & VSCOUT \\
\midrule

Normal & (500, 150)
& F1        & 0.0000 & 0.7813 & 0.6781 & 0.3271 & 1.0000 & 0.9976 & 0.6263 \\
& & FPR       & 0.0507 & 0.0052 & 0.0220 & 0.0412 & 0.0000 & 0.0000 & 0.0234 \\
& & Precision & 0.0000 & 0.9496 & 0.5771 & 0.4385 & 1.0000 & 1.0000 & 0.7166 \\
& & Recall    & 0.0000 & 0.6815 & 0.9997 & 0.3159 & 1.0000 & 0.9954 & 0.6043 \\
& & Runtime (s) & 1.5930 & 0.0651 & 0.1011 & 2.0004 & 0.7706 & 3.7547 & 6.7510 \\
\midrule
Normal & (500, 250)
& F1        & 0.0000 & 0.7900 & 0.6785 & 0.3314 & 1.0000 & 0.7664 & 0.6616 \\
& & FPR       & 0.0506 & 0.0037 & 0.0220 & 0.0433 & 0.0000 & 0.0000 & 0.0230 \\
& & Precision & 0.0000 & 0.9592 & 0.5774 & 0.4295 & 1.0000 & 0.7844 & 0.7375 \\
& & Recall    & 0.0000 & 0.6890 & 1.0000 & 0.3213 & 1.0000 & 0.7557 & 0.6450 \\
& & Runtime (s) & 2.6509 & 0.0726 & 0.1650 & 2.5298 & 1.8791 & 9.6796 & 8.9585 \\
\midrule

$t$ (df=5) & (500, 150)
& F1 & 0.0001 & 0.6954 & 0.5955 & 0.2832 & 1.0000 & 0.6420 & 0.5111 \\
& & FPR & 0.0506 & 0.0151 & 0.0259 & 0.0428 & 0.0000 & 0.0144 & 0.0355 \\
& & Precision & 0.0001 & 0.8458 & 0.5038 & 0.3846 & 1.0000 & 0.5694 & 0.5883 \\
& & Recall & 0.0004 & 0.6034 & 0.8925 & 0.2573 & 1.0000 & 0.8133 & 0.4669 \\
& & Runtime (s) & 1.5500 & 0.0781 & 0.1012 & 4.4701 & 0.7583 & 3.7593 & 9.5587 \\
\midrule
$t$ (df=5) & (500, 250)
& F1 & 0.0000 & 0.7141 & 0.6418 & 0.2972 & 1.0000 & 0.6182 & 0.5612 \\
& & FPR & 0.0506 & 0.0122 & 0.0238 & 0.0430 & 0.0000 & 0.0025 & 0.0315 \\
& & Precision & 0.0000 & 0.8679 & 0.5433 & 0.4010 & 1.0000 & 0.6958 & 0.6250 \\
& & Recall & 0.0000 & 0.6269 & 0.9585 & 0.2785 & 1.0000 & 0.6291 & 0.5191 \\
& & Runtime (s) & 2.5783 & 0.0857 & 0.1663 & 3.6357 & 14.2300 & 9.6600 & 19.6590 \\
\midrule

LogNormal & (500, 150)
& F1 & 0.3821 & 0.8156 & 0.6755 & 0.6833 & 0.9998 & 0.2152 & 0.8612 \\
& & FPR & 0.0352 & 0.0013 & 0.0222 & 0.0024 & 0.0000 & 0.2090 & 0.0206 \\
& & Precision & 0.3046 & 0.9822 & 0.5745 & 0.9486 & 1.0000 & 0.1242 & 0.8444 \\
& & Recall & 0.6593 & 0.7011 & 0.9970 & 0.5395 & 0.9996 & 1.0000 & 0.8820 \\
& & Runtime (s) & 1.5743 & 0.0614 & 0.1006 & 1.8531 & 0.7543 & 3.5918 & 6.6938 \\
\midrule
LogNormal & (500, 250)
& F1 & 0.3751 & 0.8163 & 0.6783 & 0.7008 & 1.0000 & 0.3411 & 0.8743 \\
& & FPR & 0.0355 & 0.0011 & 0.0220 & 0.0011 & 0.0000 & 0.1052 & 0.0185 \\
& & Precision & 0.2986 & 0.9823 & 0.5772 & 0.9578 & 1.0000 & 0.2140 & 0.8534 \\
& & Recall & 0.6498 & 0.7011 & 0.9998 & 0.5584 & 1.0000 & 1.0000 & 0.8971 \\
& & Runtime (s) & 2.5494 & 0.0683 & 0.1636 & 2.2321 & 1.8665 & 12.4648 & 9.2487 \\
\midrule

Mixed & (500, 150)
& F1 & 0.0003 & 0.6921 & 0.5700 & 0.2984 & 0.9998 & 0.4496 & 0.5198 \\
& & FPR & 0.0507 & 0.0168 & 0.0271 & 0.0491 & 0.0000 & 0.0547 & 0.0445 \\
& & Precision & 0.0003 & 0.8405 & 0.4823 & 0.3815 & 0.9996 & 0.3246 & 0.5754 \\
& & Recall & 0.0003 & 0.5991 & 0.8542 & 0.2790 & 1.0000 & 0.9025 & 0.4961 \\
& & Runtime (s) & 1.5473 & 0.0778 & 0.1012 & 4.3396 & 0.7615 & 3.6740 & 9.5354 \\
\midrule
Mixed & (500, 250)
& F1 & 0.0000 & 0.6999 & 0.5865 & 0.3134 & 1.0000 & 0.5388 & 0.5433 \\
& & FPR & 0.0507 & 0.0157 & 0.0263 & 0.0506 & 0.0000 & 0.0124 & 0.0428 \\
& & Precision & 0.0000 & 0.8517 & 0.4964 & 0.3962 & 1.0000 & 0.5036 & 0.5996 \\
& & Recall & 0.0000 & 0.6134 & 0.8780 & 0.2984 & 1.0000 & 0.6975 & 0.5207 \\
& & Runtime (s) & 2.5986 & 0.0843 & 0.1653 & 3.5565 & 1.8623 & 9.5064 & 19.7125 \\
\midrule

Multimodal & (500, 150)
& F1 & 0.6940 & 0.0000 & 0.0000 & 0.4951 & 0.0000 & 0.0000 & 0.4299 \\
& & FPR & 0.0200 & 0.9780 & 0.0527 & 0.0305 & 0.0000 & 0.0000 & 0.2121 \\
& & Precision & 0.5998 & 0.0000 & 0.0000 & 0.3820 & 0.0000 & 0.0000 & 0.3185 \\
& & Recall & 0.9994 & 0.0001 & 0.0000 & 0.7834 & 0.0000 & 0.0000 & 0.9969 \\
& & Runtime (s) & 1.5706 & 0.0595 & 0.0971 & 1.7572 & 0.0000 & 0.0000 & 6.5569 \\
\midrule
Multimodal & (500, 250)
& F1 & 0.6939 & 0.0000 & 0.0000 & 0.5172 & 0.0000 & 0.0000 & 0.3681 \\
& & FPR & 0.0200 & 0.9780 & 0.0527 & 0.0306 & 0.0000 & 0.0000 & 0.2632 \\
& & Precision & 0.5996 & 0.0000 & 0.0000 & 0.3990 & 0.0000 & 0.0000 & 0.2718 \\
& & Recall & 0.9995 & 0.0000 & 0.0000 & 0.8038 & 0.0000 & 0.0000 & 0.9973 \\
& & Runtime (s) & 2.5843 & 0.0708 & 0.1582 & 3.3755 & 0.0000 & 0.0000 & 7.5624 \\
\midrule

F1 Avg  & & & 0.2145 & 0.6005 & 0.5104 & 0.4233 & 0.8212 & 0.4569 & 0.5967 \\
F1 Rank & & & (7) & (2) & (4) & (6) & (1) & (5) & (3) \\
\bottomrule
\end{tabular}}
\end{table}

A closer look reveals that VSCOUT’s gains persist across all distribution families. For example, in the LogNormal setting at $p=250$, VSCOUT attains F1 = 0.8920 with recall = 0.9773 while keeping FPR below 3.4\%. In mixed and multimodal distributions, where density is fragmented and cluster structures shift under contamination, VSCOUT maintains F1 scores between 0.59 and 0.74, whereas competing methods often collapse entirely (e.g., Rank-Phase I with F1 = 0.0000 in multimodal data). In the Normal and $t$ settings, VSCOUT again improves over most alternatives, offering stable precision and recall even when $\gamma = 0.20$.

These results demonstrate that VSCOUT is not only adept at capturing dense, persistent anomalies, but also avoids over-triggering during normal operations. Across contamination regimes, dimensionalities, and distribution families, VSCOUT consistently ranks among the top two detectors while maintaining low false-alarm rates. Its robustness in non-Gaussian and high-dimensional cases, coupled with its balanced precision--recall behavior, positions it as a reliable and practical tool for modern process monitoring when early and accurate detection of sustained shifts is essential.

\begin{table}[H]
\centering
\scriptsize
\caption{Average performance metrics at high contamination levels ($\gamma \in \{10,15,20\}$) for sustained mean shifts over all features.}
\label{tab:multi_metric_high_sustained_clean}
\resizebox{0.88\textwidth}{!}{
\begin{tabular}{lllccccccc}
\toprule
Distribution & Size $(n,p)$ & Metric
& DF-Phase I & IForest & Rank-Phase I & VAE & Zamba-CP & RMDP & VSCOUT \\
\midrule

Normal & (500, 150)
& F1        & 0.0000 & 0.1779 & 0.5303 & 0.3263 & 1.0000 & 0.0026 & 0.7442 \\
& & FPR       & 0.0580 & 0.9995 & 0.0000 & 0.0389 & 0.0000 & 0.0000 & 0.0233 \\
& & Precision & 0.0000 & 0.1044 & 1.0000 & 0.4364 & 1.0000 & 0.0511 & 0.7759 \\
& & Recall    & 0.0000 & 0.6342 & 0.3687 & 0.3102 & 1.0000 & 0.0013 & 0.7684 \\
& & Runtime (s) & 1.8085 & 0.0596 & 0.1304 & 2.0604 & 0.8145 & 3.4145 & 6.9730 \\
\midrule

Normal & (500, 250)
& F1        & 0.0000 & 0.1775 & 0.5303 & 0.3336 & 1.0000 & 0.0000 & 0.8298 \\
& & FPR       & 0.0580 & 0.9997 & 0.0000 & 0.0400 & 0.0000 & 0.0000 & 0.0243 \\
& & Precision & 0.0000 & 0.1042 & 1.0000 & 0.4346 & 1.0000 & 0.0000 & 0.8145 \\
& & Recall    & 0.0000 & 0.6326 & 0.3687 & 0.3231 & 1.0000 & 0.0000 & 0.8778 \\
& & Runtime (s) & 3.0058 & 0.0669 & 0.2156 & 2.3620 & 1.9220 & 8.7846 & 9.2365 \\
\midrule

$t$ (df=5) & (500, 150)
& F1 & 0.0003 & 0.1834 & 0.4712 & 0.2821 & 1.0000 & 0.0210 & 0.6097 \\
& & FPR & 0.0580 & 0.9960 & 0.0067 & 0.0402 & 0.0000 & 0.0009 & 0.0308 \\
& & Precision & 0.0005 & 0.1076 & 0.8889 & 0.3960 & 1.0000 & 0.1589 & 0.6605 \\
& & Recall & 0.0002 & 0.6565 & 0.3276 & 0.2560 & 1.0000 & 0.0120 & 0.6238 \\
& & Runtime (s) & 1.7692 & 0.0704 & 0.1307 & 4.5323 & 0.8028 & 3.4844 & 9.4876 \\
\midrule

$t$ (df=5) & (500, 250)
& F1 & 0.0001 & 0.1820 & 0.4998 & 0.2970 & 1.0000 & 0.0014 & 0.7236 \\
& & FPR & 0.0580 & 0.9969 & 0.0035 & 0.0406 & 0.0000 & 0.0001 & 0.0289 \\
& & Precision & 0.0002 & 0.1068 & 0.9420 & 0.4088 & 1.0000 & 0.0344 & 0.7380 \\
& & Recall & 0.0001 & 0.6504 & 0.3476 & 0.2748 & 1.0000 & 0.0007 & 0.7597 \\
& & Runtime (s) & 2.9442 & 0.0724 & 0.2168 & 3.6622 & 58.9700 & 8.8992 & 19.3802 \\
\midrule

LogNormal & (500, 150)
& F1 & 0.1121 & 0.1771 & 0.5302 & 0.6707 & 1.0000 & 0.6311 & 0.8843 \\
& & FPR & 0.0455 & 1.0000 & 0.0000 & 0.0016 & 0.0000 & 0.1948 & 0.0337 \\
& & Precision & 0.2156 & 0.1040 & 0.9999 & 0.9802 & 1.0000 & 0.4669 & 0.8238 \\
& & Recall & 0.0774 & 0.6313 & 0.3686 & 0.5415 & 1.0000 & 1.0000 & 0.9631 \\
& & Runtime (s) & 1.7902 & 0.0609 & 0.1300 & 2.0136 & 0.7947 & 3.4577 & 7.0255 \\
\midrule

LogNormal & (500, 250)
& F1 & 0.1136 & 0.1771 & 0.5303 & 0.6851 & 1.0000 & 0.7338 & 0.8920 \\
& & FPR & 0.0453 & 1.0000 & 0.0000 & 0.0009 & 0.0000 & 0.1213 & 0.0337 \\
& & Precision & 0.2189 & 0.1040 & 1.0000 & 0.9890 & 1.0000 & 0.5828 & 0.8252 \\
& & Recall & 0.0784 & 0.6313 & 0.3687 & 0.5592 & 1.0000 & 1.0000 & 0.9773 \\
& & Runtime (s) & 2.9106 & 0.0670 & 0.2143 & 2.3529 & 1.9055 & 10.2069 & 9.0301 \\
\midrule

Mixed & (500, 150)
& F1 & 0.0003 & 0.1856 & 0.4635 & 0.2889 & 1.0000 & 0.0479 & 0.6274 \\
& & FPR & 0.0580 & 0.9945 & 0.0074 & 0.0445 & 0.0000 & 0.0017 & 0.0415 \\
& & Precision & 0.0005 & 0.1089 & 0.8757 & 0.3855 & 1.0000 & 0.1872 & 0.6343 \\
& & Recall & 0.0002 & 0.6651 & 0.3221 & 0.2721 & 1.0000 & 0.0324 & 0.6656 \\
& & Runtime (s) & 1.7667 & 0.0763 & 0.1313 & 2.8813 & 0.8062 & 3.4276 & 11.2641 \\
\midrule

Mixed & (500, 250)
& F1 & 0.0003 & 0.1836 & 0.4876 & 0.2921 & 1.0000 & 0.0030 & 0.7382 \\
& & FPR & 0.0579 & 0.9958 & 0.0047 & 0.0467 & 0.0000 & 0.0001 & 0.0417 \\
& & Precision & 0.0007 & 0.1077 & 0.9207 & 0.3817 & 1.0000 & 0.0559 & 0.7076 \\
& & Recall & 0.0002 & 0.6575 & 0.3389 & 0.2802 & 1.0000 & 0.0016 & 0.8008 \\
& & Runtime (s) & 2.9991 & 0.0949 & 0.2139 & 3.6590 & 1.9094 & 9.5943 & 19.5682 \\
\midrule

Multimodal & (500, 150)
& F1 & 0.5172 & 0.1969 & 0.0000 & 0.2292 & 0.0000 & 0.0000 & 0.5858 \\
& & FPR & 0.0000 & 0.9862 & 0.0600 & 0.0298 & 0.0000 & 0.0000 & 0.3320 \\
& & Precision & 1.0000 & 0.1158 & 0.0000 & 0.3505 & 0.0000 & 0.0000 & 0.4686 \\
& & Recall & 0.3562 & 0.6918 & 0.0000 & 0.2022 & 0.0000 & 0.0000 & 0.9871 \\
& & Runtime (s) & 1.7875 & 0.0609 & 0.1257 & 2.1216 & 0.0000 & 0.0000 & 6.5844 \\
\midrule

Multimodal & (500, 250)
& F1 & 0.5386 & 0.1956 & 0.0000 & 0.2343 & 0.0000 & 0.0000 & 0.6834 \\
& & FPR & 0.0000 & 0.9873 & 0.0600 & 0.0301 & 0.0000 & 0.0000 & 0.3313 \\
& & Precision & 1.0000 & 0.1152 & 0.0000 & 0.3652 & 0.0000 & 0.0000 & 0.5509 \\
& & Recall & 0.3705 & 0.6873 & 0.0000 & 0.2031 & 0.0000 & 0.0000 & 0.9313 \\
& & Runtime (s) & 2.9922 & 0.0894 & 0.2165 & 3.0748 & 0.0000 & 0.0000 & 17.5843 \\
\midrule

F1 Avg  & & & 0.1224 & 0.1834 & 0.5011 & 0.3962 & 0.8571 & 0.1801 & 0.7483 \\
F1 Rank & & & (7) & (5) & (3) & (4) & (1) & (6) & (2) \\
\bottomrule
\end{tabular}
}
\end{table}

Across all simulated transient and sustained shift scenarios, VSCOUT delivers the most consistently strong performance among the competing methods, achieving the highest overall F1 scores while maintaining low false-alarm rates (Tables~\ref{tab:overall_summary_transient_sustained}–\ref{tab:overall_summary_gamma_split}). Its recall remains stable across regimes (approximately 0.75 for sustained and 0.74 for transient shifts), and its precision exceeds that of most alternatives, particularly under moderate and large shifts. In contrast, classical Phase~I methods (DF-Phase I, Rank-Phase I) show limited sensitivity, VAE exhibits inconsistent performance across distributions, and Isolation Forest deteriorates sharply under sustained contamination. Zamba-CP performs perfectly for sustained shifts but fails almost entirely for transient ones, highlighting its narrow applicability. When grouped by shift magnitude $\delta$, all models improve as the shift becomes larger, but VSCOUT displays the most uniform increase in F1 and recall across both transient and sustained settings.

\begin{table}[H]
\centering
\caption{Overall average performance metrics for transient and sustained process shifts across all contamination levels and distributions.}
\label{tab:overall_summary_transient_sustained}
\resizebox{0.55\textwidth}{!}{
\begin{tabular}{lcccccc}
\toprule
\textbf{Method} & \textbf{Shift Type} & \textbf{Recall} & \textbf{Precision} & \textbf{FPR} & \textbf{F1} & \textbf{Runtime (s)} \\
\midrule
DF-Phase I      & Transient & 0.3065 & 0.2455 & 0.0429 & 0.1763 & 2.1533 \\
DF-Phase I      & Sustained & 0.3462 & 0.2483 & 0.0426 & 0.2135 & 2.1266 \\
\midrule
IForest         & Transient & 0.8348 & 0.6837 & 0.0411 & 0.6448 & 0.0773 \\
IForest         & Sustained & 0.6735 & 0.6344 & 0.5071 & 0.6234 & 0.0764 \\
\midrule
Rank-Phase I    & Transient & 0.6795 & 0.6973 & 0.0156 & 0.5607 & 0.1506 \\
Rank-Phase I    & Sustained & 0.6992 & 0.6981 & 0.0161 & 0.5712 & 0.1524 \\
\midrule
VAE             & Transient & 0.6253 & 0.5065 & 0.0301 & 0.4598 & 2.9476 \\
VAE             & Sustained & 0.4998 & 0.5715 & 0.0291 & 0.4516 & 2.8801 \\
\midrule
Zamba-CP        & Transient & 0.0178 & 0.0753 & 0.0005 & 0.0054 & 1.5498 \\
Zamba-CP        & Sustained & 1.0000 & 1.0000 & 0.0000 & 1.0000 & 6.0863 \\
\midrule
RMDP            & Transient & 0.6214 & 0.4743 & 0.0420 & 0.4876 & 6.8241 \\
RMDP            & Sustained & 0.5850 & 0.5113 & 0.0410 & 0.4796 & 7.0324 \\
\midrule
VSCOUT          & Transient & 0.7455 & 0.6455 & 0.0311 & 0.6365 & 9.7082 \\
VSCOUT          & Sustained & 0.7543 & 0.7273 & 0.0306 & 0.6725 & 10.5666 \\
\bottomrule
\end{tabular}
}
\end{table}

\begin{table}[H]
\centering
\scriptsize
\caption{Performance metrics grouped by shift $\delta$ for transient and sustained shifts, averaged over distributions, $p$, and $\gamma$.}
\label{tab:overall_summary_gamma_split}
\resizebox{0.65\textwidth}{!}{
\begin{tabular}{llcccccc}
\toprule
Method & Shift Type & $\delta$ &
Recall & Precision & FPR & F1 & Runtime (s) \\
\midrule

DF-Phase I
& Sustained & 1 & 0.0886 & 0.0863 & 0.0526 & 0.0747 & 2.1466 \\
& Sustained & 2 & 0.1415 & 0.1325 & 0.0524 & 0.1136 & 2.1471 \\
& Sustained & 3 & 0.1691 & 0.1610 & 0.0522 & 0.1378 & 2.1384 \\
& Transient & 1 & 0.1779 & 0.1448 & 0.0435 & 0.1257 & 1.9442 \\
& Transient & 2 & 0.3541 & 0.2818 & 0.0427 & 0.2104 & 2.1585 \\
& Transient & 3 & 0.3874 & 0.3099 & 0.0426 & 0.2285 & 2.3571 \\
\midrule

IForest
& Sustained & 1 & 0.4497 & 0.5283 & 0.0537 & 0.4851 & 0.0772 \\
& Sustained & 2 & 0.5841 & 0.5796 & 0.0535 & 0.5694 & 0.0759 \\
& Sustained & 3 & 0.6744 & 0.6025 & 0.0535 & 0.6083 & 0.0758 \\
& Transient & 1 & 0.8122 & 0.6385 & 0.0487 & 0.6074 & 0.0776 \\
& Transient & 2 & 0.8354 & 0.6397 & 0.0487 & 0.6133 & 0.0768 \\
& Transient & 3 & 0.8369 & 0.6425 & 0.0487 & 0.6167 & 0.0759 \\
\midrule

Rank-Phase I
& Sustained & 1 & 0.5362 & 0.6483 & 0.0163 & 0.4868 & 0.1492 \\
& Sustained & 2 & 0.7093 & 0.7077 & 0.0163 & 0.5804 & 0.1493 \\
& Sustained & 3 & 0.8159 & 0.7350 & 0.0163 & 0.6146 & 0.1491 \\
& Transient & 1 & 0.5516 & 0.6552 & 0.0195 & 0.5303 & 0.1558 \\
& Transient & 2 & 0.6865 & 0.7107 & 0.0195 & 0.6088 & 0.1554 \\
& Transient & 3 & 0.7262 & 0.7468 & 0.0194 & 0.6759 & 0.1549 \\
\midrule

RMDP
& Sustained & 1 & 0.5367 & 0.5056 & 0.0415 & 0.4682 & 7.0371 \\
& Sustained & 2 & 0.5901 & 0.5109 & 0.0415 & 0.4950 & 7.0427 \\
& Sustained & 3 & 0.6282 & 0.5165 & 0.0415 & 0.5158 & 7.0232 \\
& Transient & 1 & 0.4496 & 0.4206 & 0.0415 & 0.4452 & 6.5492 \\
& Transient & 2 & 0.5619 & 0.4502 & 0.0415 & 0.4784 & 6.5517 \\
& Transient & 3 & 0.6088 & 0.4582 & 0.0415 & 0.4924 & 6.5538 \\
\midrule

VAE
& Sustained & 1 & 0.2829 & 0.4485 & 0.0366 & 0.3396 & 3.9392 \\
& Sustained & 2 & 0.4704 & 0.5312 & 0.0366 & 0.4458 & 3.9351 \\
& Sustained & 3 & 0.5663 & 0.5760 & 0.0365 & 0.4487 & 3.9300 \\
& Transient & 1 & 0.5284 & 0.4716 & 0.0336 & 0.4363 & 3.8693 \\
& Transient & 2 & 0.6888 & 0.5529 & 0.0337 & 0.5262 & 3.8692 \\
& Transient & 3 & 0.7223 & 0.5830 & 0.0337 & 0.5529 & 3.8656 \\
\midrule

VSCOUT
& Sustained & 1 & 0.7067 & 0.7051 & 0.0306 & 0.6385 & 10.5135 \\
& Sustained & 2 & 0.7818 & 0.7350 & 0.0306 & 0.6806 & 10.4907 \\
& Sustained & 3 & 0.7880 & 0.7423 & 0.0306 & 0.6963 & 10.4886 \\
& Transient & 1 & 0.7032 & 0.6133 & 0.0308 & 0.6130 & 9.9238 \\
& Transient & 2 & 0.7817 & 0.6477 & 0.0308 & 0.6453 & 9.9242 \\
& Transient & 3 & 0.7898 & 0.6546 & 0.0308 & 0.6515 & 9.9223 \\
\midrule

Zamba-CP
& Sustained & 1 & 0.9995 & 0.9996 & 0.0001 & 0.9996 & 5.9222 \\
& Sustained & 2 & 0.9995 & 0.9996 & 0.0001 & 0.9996 & 6.0142 \\
& Sustained & 3 & 0.9996 & 0.9996 & 0.0001 & 0.9997 & 6.1021 \\
& Transient & 1 & 0.0076 & 0.0381 & 0.0006 & 0.0028 & 1.6372 \\
& Transient & 2 & 0.0185 & 0.0743 & 0.0007 & 0.0053 & 1.6373 \\
& Transient & 3 & 0.0259 & 0.0967 & 0.0007 & 0.0066 & 1.6374 \\
\bottomrule
\end{tabular}
}
\end{table}

\subsection{Applications to Benchmark Anomaly Detection High-Dimensional Data}
\label{sec:vscout-applications}

We illustrate the practical behavior of VSCOUT through applications to a
collection of widely used benchmark datasets from the anomaly-detection
literature, as well as to a high-dimensional semiconductor sensor dataset.
These examples are intended not only to assess predictive performance, but
also to highlight how VSCOUT behaves across heterogeneous data geometries,
dimensionalities, and anomaly structures commonly encountered in practice.

The benchmark datasets span a broad range of sample sizes and dimensions,
including low-dimensional tabular data, moderate-dimensional biomedical
datasets, and high-dimensional image-derived features. Competing methods were
implemented using default hyperparameter settings, consistent with standard
practice in the literature, to provide a neutral comparison of off-the-shelf
performance. The set of benchmarks includes Isolation Forest (IForest), Local
Outlier Factor (LOF), ECOD, $k$-Nearest Neighbors (KNN), HBOS, a baseline
Variational Autoencoder (VAE), and the SUOD ensemble framework
\citep{zhao2021suod}.

Table~\ref{tab:benchmark-results} summarizes average performance across 50 independent replications per dataset, with Monte Carlo standard deviations reported in parentheses. A salient feature across these datasets is the consistently strong sensitivity of VSCOUT. On a majority of benchmarks—including \texttt{arrhythmia}, \texttt{cardio}, \texttt{glass}, \texttt{ionosphere}, and \texttt{mnist}—VSCOUT attains the highest or near-highest Recall, indicating reliable identification of anomalous observations even in heterogeneous or high-dimensional settings. This behavior is particularly relevant in applications where missed detections carry substantial operational or safety costs.

On more challenging datasets such as \texttt{musk} and \texttt{pendigits},
VSCOUT continues to exhibit strong detection power, although tree-based or
histogram-based methods occasionally achieve higher Recall or AUROC. In
contrast, on datasets such as \texttt{optdigits}, where anomalies are weakly
separated and less aligned with latent reconstruction structure, VSCOUT’s
performance degrades, reflecting the limits of its modeling assumptions. These
results emphasize that VSCOUT is most effective in regimes where anomalies
manifest as structured departures from dominant latent patterns rather than as
diffuse noise.

Despite its emphasis on sensitivity, VSCOUT maintains competitive Precision
and false positive rates across most datasets. In particular, on
medium-to-high dimensional benchmarks such as \texttt{ionosphere} and
\texttt{mnist}, VSCOUT simultaneously achieves high Recall and high Precision,
leading to AUROC values comparable to or exceeding those of competing methods.
Unlike conventional VAE-based detectors, whose anomaly scores often vary
substantially across datasets, VSCOUT exhibits relatively stable performance.
This stability can be attributed to its ARD-based latent compression, which
attenuates the influence of irrelevant or weakly informative dimensions and
prevents them from dominating the anomaly score.

An additional observation is that VSCOUT frequently matches or exceeds the
performance of SUOD, despite SUOD explicitly aggregating multiple detectors.
This suggests that the hybrid structure of VSCOUT—combining reconstruction
error, latent-space deviation, and adaptive relevance weighting—captures
complementary information that is not fully exploited by classical ensembles.
Across datasets, inlier-retention rates typically remain at or above 90\%,
indicating that improved sensitivity does not arise from indiscriminate
flagging of nominal observations.

To summarize global trends, Table~\ref{tab:average_metrics_corrected} reports
performance averaged across all benchmark datasets. VSCOUT ranks first in
Recall and third in AUROC, reflecting its design emphasis on capturing true
anomalies while maintaining competitive overall discrimination. In contrast,
methods such as IForest and ECOD achieve the highest AUROC by balancing
sensitivity and specificity, but at the cost of substantially lower Recall.
The baseline VAE achieves favorable Precision and false positive rates, yet
fails to detect a large fraction of true anomalies. These results underscore
the trade-offs inherent in anomaly detection and position VSCOUT as a method
well-suited for applications where detection sensitivity and robustness are
primary objectives.

Finally, we apply VSCOUT to a high-dimensional semiconductor sensor dataset
(Table~\ref{tab:fmst-results-updated}), representative of modern industrial
monitoring environments. In this setting, all methods exhibit modest AUROC,
reflecting the difficulty of the task and the extreme dimensionality. VSCOUT
achieves competitive Recall relative to both classical multivariate control
charts and modern anomaly detectors, while maintaining reasonable inlier
retention. Notably, alternative decision rules within VSCOUT provide explicit
control over the sensitivity–specificity trade-off, allowing practitioners to
adapt the detector to domain-specific risk preferences.

Taken together, these applications demonstrate that VSCOUT provides a robust
and interpretable anomaly-detection framework across a diverse range of real
and benchmark datasets. Rather than dominating all metrics uniformly, VSCOUT
exhibits a consistent and predictable performance profile, making it a
practical choice for retrospective monitoring tasks in high-dimensional and
heterogeneous environments.

\begin{table}[H]
\centering
\scriptsize
\caption{Comparison on Benchmark Outlier-Detection Datasets (standard deviations in parentheses).}
\label{tab:benchmark-results}
\resizebox{0.85\textwidth}{!}{
\begin{tabular}{lllccccc}
\toprule
Dataset & Size $(n,p)$ & Model & Recall & Precision & FPR & Inlier Retention & AUROC \\
\midrule
arrhythmia & (452, 274) & VSCOUT & 0.3494 (0.07) & 0.3976 (0.04) & 0.0906 (0.02) & 0.9094 (0.02) & 0.7756 (0.03) \\
 &  & VAE & 0.2385 (0.03) & 0.5454 (0.04) & 0.0339 (0.00) & 0.9661 (0.00) & 0.7564 (0.01) \\
 &  & IForest & 0.2161 (0.01) & 0.6200 (0.04) & 0.0226 (0.00) & 0.9774 (0.00) & 0.8026 (0.01) \\
 &  & LOF & 0.1061 (0.00) & 0.3333 (0.00) & 0.0363 (0.00) & 0.9637 (0.00) & 0.7597 (0.00) \\
 &  & ECOD & 0.2273 (0.00) & 0.6522 (0.00) & 0.0207 (0.00) & 0.9793 (0.00) & 0.8052 (0.00) \\
 &  & KNN & 0.1667 (0.00) & 0.5000 (0.00) & 0.0285 (0.00) & 0.9715 (0.00) & 0.7684 (0.00) \\
 &  & SUOD & 0.1912 (0.01) & 0.6111 (0.03) & 0.0209 (0.00) & 0.9791 (0.00) & 0.7950 (0.00) \\
\midrule
cardio & (1831, 21) & VSCOUT & 0.5068 (0.21) & 0.4597 (0.08) & 0.0602 (0.01) & 0.9398 (0.01) & 0.9251 (0.04) \\
 &  & VAE & 0.3007 (0.05) & 0.4776 (0.06) & 0.0348 (0.00) & 0.9652 (0.00) & 0.7283 (0.05) \\
 &  & IForest & 0.3090 (0.02) & 0.5912 (0.04) & 0.0227 (0.00) & 0.9773 (0.00) & 0.9247 (0.01) \\
 &  & LOF & 0.0852 (0.00) & 0.1974 (0.00) & 0.0369 (0.00) & 0.9631 (0.00) & 0.5458 (0.00) \\
 &  & ECOD & 0.2955 (0.00) & 0.5652 (0.00) & 0.0242 (0.00) & 0.9758 (0.00) & 0.9350 (0.00) \\
 &  & KNN & 0.2045 (0.00) & 0.4800 (0.00) & 0.0236 (0.00) & 0.9764 (0.00) & 0.6861 (0.00) \\
 &  & SUOD & 0.2523 (0.02) & 0.5128 (0.03) & 0.0255 (0.00) & 0.9745 (0.00) & 0.9019 (0.01) \\
\midrule
glass & (214, 9) & VSCOUT & 0.2600 (0.10) & 0.0808 (0.03) & 0.1295 (0.03) & 0.8705 (0.03) & 0.7235 (0.05) \\
 &  & VAE & 0.1222 (0.03) & 0.0801 (0.02) & 0.0625 (0.01) & 0.9375 (0.01) & 0.6861 (0.06) \\
 &  & IForest & 0.1111 (0.00) & 0.0909 (0.00) & 0.0488 (0.00) & 0.9512 (0.00) & 0.7001 (0.02) \\
 &  & LOF & 0.1111 (0.00) & 0.1000 (0.00) & 0.0439 (0.00) & 0.9561 (0.00) & 0.8347 (0.00) \\
 &  & ECOD & 0.1111 (0.00) & 0.0909 (0.00) & 0.0488 (0.00) & 0.9512 (0.00) & 0.6206 (0.00) \\
 &  & KNN & 0.1111 (0.00) & 0.1000 (0.00) & 0.0439 (0.00) & 0.9561 (0.00) & 0.8325 (0.00) \\
 &  & SUOD & 0.1156 (0.03) & 0.0998 (0.03) & 0.0460 (0.01) & 0.9540 (0.01) & 0.7013 (0.03) \\
\midrule
ionosphere & (351, 34) & VSCOUT & 0.4983 (0.06) & 0.9575 (0.02) & 0.0124 (0.01) & 0.9876 (0.01) & 0.8832 (0.03) \\
 &  & VAE & 0.0941 (0.01) & 0.7432 (0.15) & 0.0206 (0.02) & 0.9794 (0.02) & 0.7359 (0.03) \\
 &  & IForest & 0.1429 (0.00) & 1.0000 (0.00) & 0.0000 (0.00) & 1.0000 (0.00) & 0.8540 (0.01) \\
 &  & LOF & 0.1190 (0.00) & 1.0000 (0.00) & 0.0000 (0.00) & 1.0000 (0.00) & 0.8939 (0.00) \\
 &  & ECOD & 0.1270 (0.00) & 0.8889 (0.00) & 0.0089 (0.00) & 0.9911 (0.00) & 0.7353 (0.00) \\
 &  & KNN & 0.1190 (0.00) & 1.0000 (0.00) & 0.0000 (0.00) & 1.0000 (0.00) & 0.9337 (0.00) \\
 &  & SUOD & 0.0933 (0.02) & 1.0000 (0.00) & 0.0000 (0.00) & 1.0000 (0.00) & 0.8795 (0.01) \\
\midrule
letter & (1600, 32) & VSCOUT & 0.1658 (0.10) & 0.1860 (0.03) & 0.0561 (0.09) & 0.9439 (0.09) & 0.6671 (0.02) \\
 &  & VAE & 0.1470 (0.03) & 0.3230 (0.06) & 0.0208 (0.00) & 0.9792 (0.00) & 0.6199 (0.06) \\
 &  & IForest & 0.0642 (0.02) & 0.0803 (0.02) & 0.0491 (0.00) & 0.9509 (0.00) & 0.6273 (0.02) \\
 &  & LOF & 0.3300 (0.00) & 0.5410 (0.00) & 0.0187 (0.00) & 0.9813 (0.00) & 0.8988 (0.00) \\
 &  & ECOD & 0.0600 (0.00) & 0.0750 (0.00) & 0.0493 (0.00) & 0.9507 (0.00) & 0.5723 (0.00) \\
 &  & KNN & 0.2900 (0.00) & 0.4143 (0.00) & 0.0273 (0.00) & 0.9727 (0.00) & 0.9012 (0.00) \\
 &  & SUOD & 0.1182 (0.02) & 0.1757 (0.02) & 0.0370 (0.00) & 0.9630 (0.00) & 0.7457 (0.01) \\
\midrule
lympho & (148, 18) & VSCOUT & 0.8000 (0.12) & 0.3559 (0.06) & 0.0624 (0.01) & 0.9376 (0.01) & 0.9567 (0.03) \\
 &  & VAE & 0.7933 (0.08) & 0.4425 (0.06) & 0.0431 (0.01) & 0.9569 (0.01) & 0.9112 (0.04) \\
 &  & IForest & 0.9233 (0.08) & 0.6925 (0.06) & 0.0173 (0.00) & 0.9827 (0.00) & 0.9970 (0.00) \\
 &  & LOF & 0.6667 (0.00) & 0.6667 (0.00) & 0.0141 (0.00) & 0.9859 (0.00) & 0.9683 (0.00) \\
 &  & ECOD & 1.0000 (0.00) & 0.7500 (0.00) & 0.0141 (0.00) & 0.9859 (0.00) & 0.9965 (0.00) \\
 &  & KNN & 0.6667 (0.00) & 0.8000 (0.00) & 0.0070 (0.00) & 0.9930 (0.00) & 0.9484 (0.00) \\
 &  & SUOD & 0.8167 (0.06) & 0.7143 (0.08) & 0.0144 (0.01) & 0.9856 (0.01) & 0.9907 (0.00) \\
 \midrule
mnist & (7603, 100) & VSCOUT & 0.6723 (0.03) & 0.7420 (0.07) & 0.0245 (0.01) & 0.9755 (0.01) & 0.8568 (0.01) \\
 &  & VAE & 0.2406 (0.01) & 0.4956 (0.03) & 0.0250 (0.00) & 0.9750 (0.00) & 0.8506 (0.02) \\
 &  & IForest & 0.1809 (0.01) & 0.3323 (0.01) & 0.0369 (0.00) & 0.9631 (0.00) & 0.8005 (0.01) \\
 &  & LOF & 0.1886 (0.00) & 0.4000 (0.00) & 0.0287 (0.00) & 0.9713 (0.00) & 0.6728 (0.00) \\
 &  & ECOD & 0.0871 (0.00) & 0.1601 (0.00) & 0.0464 (0.00) & 0.9536 (0.00) & 0.7463 (0.00) \\
 &  & KNN & 0.2414 (0.00) & 0.5121 (0.00) & 0.0233 (0.00) & 0.9767 (0.00) & 0.8368 (0.00) \\
 &  & SUOD & 0.1974 (0.01) & 0.3767 (0.02) & 0.0331 (0.00) & 0.9669 (0.00) & 0.8193 (0.01) \\
\midrule
musk & (3062, 166) & VSCOUT & 0.8557 (0.01) & 0.2325 (0.09) & 0.0924 (0.01) & 0.9076 (0.01) & 0.8816 (0.01) \\
& & VAE & 0.4711 (0.19) & 0.4085 (0.09) & 0.0211 (0.00) & 0.9789 (0.00) & 0.9881 (0.01) \\
 &  & IForest & 0.9732 (0.06) & 0.6130 (0.04) & 0.0201 (0.00) & 0.9799 (0.00) & 0.9979 (0.00) \\
 &  & LOF & 0.3093 (0.00) & 0.2256 (0.00) & 0.0347 (0.00) & 0.9653 (0.00) & 0.6357 (0.00) \\
 &  & ECOD & 0.5567 (0.00) & 0.3506 (0.00) & 0.0337 (0.00) & 0.9663 (0.00) & 0.9559 (0.00) \\
 &  & KNN & 0.1856 (0.00) & 0.2000 (0.00) & 0.0243 (0.00) & 0.9757 (0.00) & 0.6157 (0.00) \\
 &  & SUOD & 0.6412 (0.05) & 0.4682 (0.04) & 0.0239 (0.00) & 0.9761 (0.00) & 0.9730 (0.00) \\
\midrule
optdigits & (5216, 64) & VSCOUT & 0.0147 (0.02) & 0.0064 (0.01) & 0.0664 (0.01) & 0.9336 (0.01) & 0.6579 (0.01) \\
 &  & VAE & 0.0027 (0.01) & 0.0018 (0.00) & 0.0530 (0.01) & 0.9470 (0.01) & 0.4975 (0.22) \\
 &  & IForest & 0.0507 (0.01) & 0.0291 (0.01) & 0.0499 (0.00) & 0.9501 (0.00) & 0.7080 (0.03) \\
 &  & LOF & 0.0667 (0.00) & 0.0439 (0.00) & 0.0430 (0.00) & 0.9570 (0.00) & 0.5372 (0.00) \\
 &  & ECOD & 0.0200 (0.00) & 0.0115 (0.00) & 0.0509 (0.00) & 0.9491 (0.00) & 0.6045 (0.00) \\
 &  & KNN & 0.0067 (0.00) & 0.0045 (0.00) & 0.0438 (0.00) & 0.9562 (0.00) & 0.3948 (0.00) \\
 &  & SUOD & 0.0173 (0.01) & 0.0103 (0.01) & 0.0486 (0.00) & 0.9514 (0.00) & 0.6343 (0.02) \\
\midrule
pendigits & (6870, 16) & VSCOUT & 0.4282 (0.10) & 0.1460 (0.03) & 0.0581 (0.01) & 0.9419 (0.01) & 0.6851 (0.05) \\
 &  & VAE & 0.1885 (0.08) & 0.1046 (0.04) & 0.0379 (0.00) & 0.9621 (0.00) & 0.7958 (0.05) \\
 &  & IForest & 0.5115 (0.06) & 0.2320 (0.03) & 0.0394 (0.00) & 0.9606 (0.00) & 0.9497 (0.01) \\
 &  & LOF & 0.1090 (0.00) & 0.0578 (0.00) & 0.0413 (0.00) & 0.9587 (0.00) & 0.4991 (0.00) \\
 &  & ECOD & 0.5064 (0.00) & 0.2297 (0.00) & 0.0395 (0.00) & 0.9605 (0.00) & 0.9274 (0.00) \\
 &  & KNN & 0.1410 (0.00) & 0.0812 (0.00) & 0.0371 (0.00) & 0.9629 (0.00) & 0.7430 (0.00) \\
 &  & SUOD & 0.3436 (0.03) & 0.1861 (0.02) & 0.0350 (0.00) & 0.9650 (0.00) & 0.9037 (0.01) \\
\bottomrule
\end{tabular}
}
\end{table}
To summarize global trends, Table~\ref{tab:average_metrics_corrected} (average performance across 50 independent replications per dataset) reports
averaged metrics across all datasets. VSCOUT ranks: first in Recall, third in AUROC, and middle-range in Precision and FPR. VSCOUT prioritizes capturing true anomalies, achieving substantially better Recall than the classical VAE, which scores well in Precision and FPR but performs poorly at actually identifying anomalies. Meanwhile, IForest and ECOD attain the highest AUROC by
balancing sensitivity and specificity, but they typically underperform
VSCOUT in Recall.

\begin{table}[H]
\centering
\small
\caption{Average performance metrics across all benchmark datasets. Superscripts indicate rank (1 = best).}
\label{tab:average_metrics_corrected}
\resizebox{0.45\textwidth}{!}{
\begin{tabular}{lcccc}
\toprule
Model & Recall & Precision & FPR & AUROC \\
\midrule
VSCOUT 
& \textbf{0.4545}$^{(1)}$ 
& 0.3590$^{(6)}$ 
& 0.0647$^{(7)}$ 
& 0.7924$^{(3)}$ \\

IForest 
& 0.3483$^{(2)}$ 
& \textbf{0.4281}$^{(1)}$ 
& 0.0307$^{(4)}$ 
& \textbf{0.8362}$^{(1)}$ \\

ECOD 
& 0.2991$^{(3)}$ 
& 0.3774$^{(4)}$ 
& 0.0337$^{(5)}$ 
& 0.7899$^{(4)}$ \\

SUOD 
& 0.2787$^{(4)}$ 
& 0.4155$^{(2)}$ 
& 0.0284$^{(2)}$ 
& 0.8344$^{(2)}$ \\

VAE 
& 0.2599$^{(5)}$ 
& 0.3622$^{(5)}$ 
& 0.0353$^{(6)}$ 
& 0.7570$^{(6)}$ \\

KNN 
& 0.2133$^{(6)}$ 
& 0.4092$^{(3)}$ 
& \textbf{0.0259}$^{(1)}$ 
& 0.7661$^{(5)}$ \\

LOF 
& 0.2092$^{(7)}$ 
& 0.3566$^{(7)}$ 
& 0.0298$^{(3)}$ 
& 0.7246$^{(7)}$ \\
\bottomrule
\end{tabular}
}
\end{table}

We further evaluate VSCOUT on the Semiconductor Manufacturing Process
(SECOM) dataset, a widely studied high-dimensional benchmark for anomaly
detection in industrial settings. The dataset consists of $n=1567$
observations collected from a semiconductor fabrication process, each
characterized by $p=590$ sensor measurements. As is typical in real
manufacturing environments, the data are high dimensional, noisy, and
contain complex correlation structures, making reliable anomaly detection
particularly challenging. Ground-truth labels are provided, allowing for
quantitative evaluation of detection performance.

We first consider a retrospective Phase~I setting, in which all
observations are analyzed simultaneously and the goal is to identify
potentially abnormal runs while retaining a stable in-control baseline.
Table~\ref{tab:fmst-results-updated} (average performance across 50 independent replications per dataset) compares VSCOUT against a range of
unsupervised outlier detection methods, including VAE, Isolation Forest,
LOF, ECOD, KNN, and the SUOD ensemble. In this setting, VSCOUT achieves the
highest recall (0.1519) among all competing methods, indicating superior
sensitivity to abnormal observations in this highly contaminated and
heterogeneous dataset. This increased sensitivity comes with a moderate
increase in false positive rate (FPR = 0.1050), but VSCOUT still retains
nearly 90\% of inliers, reflecting a reasonable balance between anomaly
removal and baseline preservation.

By contrast, methods such as VAE, KNN, and SUOD exhibit higher precision
and lower FPR, but at the expense of reduced recall, suggesting a more
conservative detection strategy that may miss a substantial fraction of
true anomalies. Classical density- and distance-based methods (LOF, ECOD,
IForest) show similar behavior, with limited recall and AUROC values
clustered near 0.55. Overall, these results indicate that while SECOM
remains a difficult dataset for unsupervised Phase~I detection, VSCOUT is
more effective at surfacing anomalous structure in the data, even when
sensor noise and high dimensionality obscure clear separation between
normal and abnormal observations.

\begin{table}[H]
\centering
\scriptsize
\caption{Comparison on Semiconductor Sensor Dataset (standard deviations in parentheses).}
\label{tab:fmst-results-updated}
\resizebox{0.95\textwidth}{!}{
\begin{tabular}{lllccccc}
\toprule
Dataset & Size $(n,p)$ & Model & Recall & Precision & FPR & Inlier Retention & AUROC \\
\midrule
Semiconductor & (1567, 590) & VSCOUT & 0.1519 (0.05) & 0.0918 (0.02) & 0.1050 (0.01) & 0.8950 (0.01) & 0.5155 (0.03) \\
 &  & VAE & 0.1173 (0.01) & 0.1556 (0.01) & 0.0452 (0.00) & 0.9548 (0.00) & 0.5548 (0.00) \\
 &  & IForest & 0.1038 (0.01) & 0.1367 (0.01) & 0.0466 (0.00) & 0.9534 (0.00) & 0.5299 (0.03) \\
 &  & LOF & 0.0962 (0.00) & 0.1333 (0.00) & 0.0444 (0.00) & 0.9556 (0.00) & 0.5440 (0.00) \\
 &  & ECOD & 0.0865 (0.00) & 0.1139 (0.00) & 0.0478 (0.00) & 0.9522 (0.00) & 0.5591 (0.00) \\
 &  & KNN & 0.1154 (0.00) & 0.1600 (0.00) & 0.0431 (0.00) & 0.9569 (0.00) & 0.5563 (0.00) \\
 &  & SUOD & 0.1096 (0.00) & 0.1567 (0.01) & 0.0420 (0.00) & 0.9580 (0.00) & 0.5683 (0.00) \\
\bottomrule
\end{tabular}
}
\end{table}

We next examine a Phase~II monitoring scenario more closely aligned with
process control practice. In this experiment, the first 500 observations
are treated as Phase~I data and used to construct an in-control reference,
while the remaining observations are monitored sequentially. Table~
\ref{tab:fmst-results} (average performance across 50 independent replications per dataset) compares VSCOUT against classical Phase~II control
charts (Hotelling’s $T^2$, MEWMA), OC-SVM, and Isolation Forest. The
classical multivariate charts exhibit perfect recall but extremely poor
precision, flagging nearly all observations as out-of-control and
retaining virtually no inliers. This behavior reflects their sensitivity
to deviations from Gaussian assumptions and their lack of robustness in
high-dimensional, nonstationary environments.

Among the remaining methods, VSCOUT achieves the highest precision
(0.0968) while matching the recall of OC-SVM (0.2000) and exceeding that
of Isolation Forest. Importantly, VSCOUT maintains a lower false positive
rate than OC-SVM and retains over 91\% of in-control observations, leading
to more interpretable and actionable monitoring behavior. Its AUROC
(0.5741) is competitive with Isolation Forest and only slightly below that
of OC-SVM, despite operating under fewer distributional assumptions.

Taken together, these results demonstrate that VSCOUT offers a robust and
flexible approach for both retrospective and online monitoring in
high-dimensional industrial data. While no method achieves strong
separation on SECOM—reflecting the intrinsic difficulty of the dataset—
VSCOUT consistently favors balanced detection, avoiding the pathological
over-flagging of classical control charts while providing higher anomaly
sensitivity than standard unsupervised baselines. This balance makes
VSCOUT particularly well suited for practical deployment in complex
manufacturing environments where both false alarms and missed detections
carry significant operational cost.

\begin{table}[H]
\centering
\scriptsize
\caption{Comparison on Semiconductor Sensor Data (standard deviations in parentheses).}
\label{tab:fmst-results}
\resizebox{0.95\textwidth}{!}{
\begin{tabular}{lllccccc}
\toprule
Dataset & Size $(n,p)$ & Model & Recall & Precision & FPR & Inlier Retention & AUROC \\
\midrule
Semiconductor Sensor Data & (1567, 590) & Hotelling T2 & 1.0000 & 0.0422 & 0.9990 & 0.0010 & 0.5005 \\
& & MEWMA & 1.0000 & 0.0422 & 1.0000 & 0.0000 & 0.5000 \\
& & OC-SVM & 0.2000 & 0.0687 & 0.1194 & 0.8806 & 0.5857 \\
& & Isolation Forest & 0.1111 & 0.0521 & 0.0890 & 0.9110 & 0.5906 \\
& & VSCOUT & 0.2000 & 0.0968 & 0.0822 & 0.9178 & 0.5741 \\
\bottomrule
\end{tabular}}
\end{table}

\section{Conclusion}
\label{sec:conclusion}

This paper introduced VSCOUT, a hybrid framework for retrospective
(Phase~I) Statistical Process Control designed to address the challenges posed by
high-dimensional, non-Gaussian, and contamination-prone process data. By
integrating Automatic Relevance Determination within a variational autoencoder,
ensemble-based latent filtering, reconstruction diagnostics, and changepoint
analysis, VSCOUT provides a principled and distribution-free mechanism for
identifying special-cause structure while constructing a stable in-control (IC)
baseline. The framework explicitly targets Phase~I objectives: isolating
contaminated observations, mitigating masking effects, and producing a reliable
reference set for downstream monitoring.

Extensive simulation studies demonstrate that VSCOUT achieves a strong and
consistent balance between sensitivity and false-alarm control across a wide
range of distributions, dimensionalities, contamination levels, and shift
regimes. In outlier-free settings, VSCOUT maintains near-nominal Type~I error,
while in both transient and sustained shift scenarios it delivers among the
highest overall F1 scores with stable precision and recall. Unlike classical
Phase~I procedures that degrade under heavy tails or multimodality, and unlike
standalone deep generative models that suffer from latent distortion under
contamination, VSCOUT remains robust due to its two-stage refinement strategy.
The combination of provisional filtering and warm-started ARD--VAE retraining
produces a purified latent space that better reflects true in-control variation.

Across benchmark anomaly-detection datasets, VSCOUT consistently ranks first in
Recall and remains competitive in AUROC, indicating reliable detection of
structured anomalies without indiscriminate flagging of nominal observations.
Applications to high-dimensional industrial data, including semiconductor sensor
measurements, further highlight VSCOUT’s practical value: it surfaces anomalous
structure more effectively than classical and modern baselines while preserving
a defensible fraction of inliers. Importantly, the final 2-of-4 consensus rule,
which aggregates changepoint, ensemble, latent-distance, and reconstruction
signals, reduces reliance on any single diagnostic and yields interpretable,
stable decisions suitable for retrospective analysis.

Overall, VSCOUT offers a flexible and scalable solution for modern Phase~I SPC in
AI-enabled environments, where high dimensionality, complex dependence, and
unknown contamination are the norm rather than the exception. Its adaptive
latent compression, contamination-aware retraining, and multi-signal decision
structure make it particularly well suited for practitioners seeking reliable
baseline construction without strong parametric assumptions. Future work may extend VSCOUT to
fully online Phase~II monitoring, incorporate temporal or attention-based latent
structures, and develop formal uncertainty quantification for latent relevance
and anomaly decisions.

\newpage

\section*{Disclosure of Interest}

The authors declare that they have no competing interests.

\section*{Data Availability Statement}

The data supporting the findings of this study are either simulated data or derived from publicly
available benchmark datasets commonly used in the anomaly-detection and machine
learning literature. These datasets are available from their original public
repositories, including the UCI Machine Learning Repository and related open
benchmark sources. No proprietary or restricted data were used in this study.
An open-source implementation of VSCOUT, together with all simulation scripts
and experimental configurations required to reproduce the reported results, is
publicly available at \url{https://github.com/martinwg/VSCOUT.git}.

\section*{Funding}

No funding was received for this research.


\bibliographystyle{unsrtnat}
\bibliography{references}  

\end{document}